%% file: main.tex
\documentclass[10pt,twocolumn,letterpaper]{article}

\usepackage[pagenumbers]{cvpr} % To force page numbers, e.g. for an arXiv version
\usepackage{times}
\usepackage{epsfig}
\usepackage{graphicx}
\usepackage{amsmath}
\usepackage{amssymb}
\usepackage{booktabs}
\usepackage{multibib}
\usepackage{xcolor,colortbl}
\usepackage{caption}
\usepackage{bm}
\usepackage{enumitem}
\usepackage{import}
\usepackage{multirow}
\usepackage{arydshln}
\usepackage[ruled,linesnumbered]{algorithm2e}
\usepackage{rotating}
\usepackage{optidef}
\usepackage{soul}
\usepackage{calc}
\usepackage{tabularx}
\usepackage{ragged2e}
\usepackage{wrapfig}
\usepackage{diagbox}
\usepackage{capt-of}
\usepackage{fixmath}
\input{preamble}

\definecolor{cvprblue}{rgb}{0.21,0.49,0.74}
\usepackage[pagebackref,breaklinks,colorlinks,citecolor=cvprblue]{hyperref}

\title{GTR: Gaussian Splatting Tracking and Reconstruction of \\
Unknown Objects Based on Appearance and Geometric Complexity}

\author{Takuya Ikeda$^{1}$
\and
Sergey Zakharov$^{2}$
\and
Muhammad Zubair Irshad$^{2}$
\and
Istvan Balazs Opra$^{1}$
\and
Shun Iwase$^{2}$
\and
Dian Chen$^{2}$
\and
Mark Tjersland$^{2}$
\and
Robert Lee$^{1}$
\and
Alexandre Dilly$^{1}$
\and
Rares Ambrus$^{2}$
\and
Koichi Nishiwaki$^{1}$
\and
\\
$^{1}$ Woven by Toyota, Inc. \hspace{5em}
$^{2}$ Toyota Research Institute % \\
}

\begin{document}

\twocolumn[{
\renewcommand\twocolumn[1][]{#1}%
\maketitle
\begin{center}
        \includegraphics[width=1.0\linewidth]{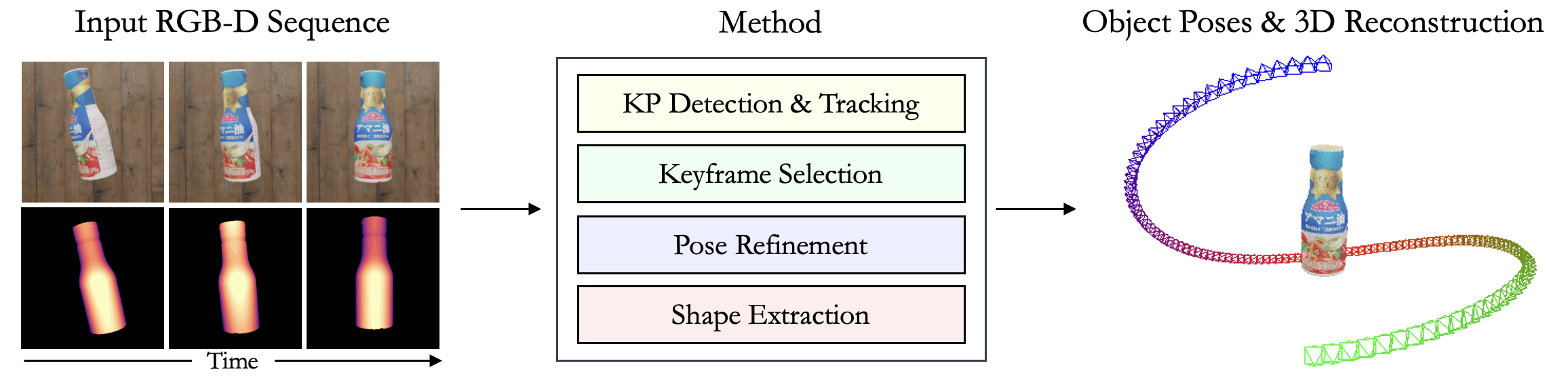}
        \captionof{figure}{We present \method, an adaptive method for 6-DoF object tracking and 3D reconstruction from monocular RGBD video. Although this shows the camera's trajectory based on the object coordinate, the fixed camera is used for the tracking of moving objects.}
        \label{fig:teaser}
\end{center}
}]

\input{sec/0_abstract}  
\input{sec/1_intro}
\input{sec/2_related}
\input{sec/3_method}
\input{sec/4_evaluation}

\input{sec/5_conclusion}

{
    \small
    \bibliographystyle{ieeenat_fullname}
    \bibliography{main}
}

\input{sec/X_suppl}


% {
%     \small
%     \bibliographystyle{ieeenat_fullname}
%     \bibliography{supp}
% }

\end{document}

%% file: preamble.tex
\newcommand{\method}{GTR\xspace}

\newcommand{\syndataset}{GTR3D Synthetic\xspace}
\newcommand{\realdataset}{GTR3D Real\xspace}

%% file: sec/0_abstract.tex
\begin{abstract}
We present a novel method for 6-DoF object tracking and high-quality 3D reconstruction from monocular RGBD video.
Existing methods, while achieving impressive results, often struggle with complex objects—particularly those exhibiting symmetry, intricate geometry or complex appearance. To bridge these gaps, we introduce an adaptive method that combines 3D Gaussian Splatting,  hybrid geometry/appearance tracking, and key frame selection to achieve robust tracking and accurate reconstructions across a diverse range of objects. Additionally, we present a benchmark covering these challenging object classes, providing high-quality annotations for evaluating both tracking and reconstruction performance. Our approach demonstrates strong capabilities in recovering high-fidelity object meshes, setting a new standard for single-sensor 3D reconstruction in open-world environments.
\end{abstract}

% {\blfootnote{*Authors contributed equally to this work.}}

%% file: sec/1_intro.tex
\section{Introduction}
\label{sec:intro}
The demand for 3D model creation, particularly of household objects, is increasing in domains such as augmented reality, robotic manipulation, and sim-to-real transfer. 
However, most current state-of-the-art reconstruction solutions rely on costly multi-camera setups, limiting accessibility. 
Achieving a 360 degree object reconstruction requires time-consuming operations, such as flipping the object to cover its bottom side, and registering its poses from before and after the manipulation.  
In this work, we aim to make object reconstruction more accessible by introducing a solution that reconstructs dynamic unknown objects from monocular RGBD video, and provide a comprehensive analysis of the components essential for successful reconstruction.

Creating open-world 3D models from a single monocular RGB-D sensor requires solving two major computer vision challenges: 6-DoF object pose tracking and 3D reconstruction~\cite{wen2023bundlesdf}. Most existing methods address these challenges independently. For example, some works focus on recovering object poses given perfect 3D model reconstructions~\cite{foundationposewen2024}, while others extend this to class-level pose estimation, though this approach remains limited to predefined object classes. Conversely, a large body of work investigates achieving high-quality reconstructions with perfect camera poses, leveraging neural representations ranging from implicit SDF-based methods to NeRF and, more recently, Gaussian Splatting representations. 
A few recent approaches tackle 6-DoF pose tracking and 3D reconstruction concurrently. BundleSDF~\cite{wen2023bundlesdf} is the first effective solution capable of jointly recovering object poses and 3D reconstructions from a monocular RGBD sequence. Although impressive, BundleSDF struggles with axis-symmetric objects and may not capture all high-frequency geometric details. 

We argue that current benchmarks fall short of providing thorough evaluation of joint pose estimation and object reconstruction on object geometries commonly encountered in household environments. Specifically, they lack sufficient testing for axis-symmetric objects, such as bottles. They also do not provide benchmarks for objects with diverse appearances and complex geometries, or performance assessments based on variable depth quality. Additionally, there is no comprehensive evaluation dataset that includes a complete view of objects spanning these challenging categories.

To address these gaps, we introduce a novel method for object-centric tracking and reconstruction of unknown objects. Our approach combines pose graph optimization with generalized point tracking, and 3D Gaussian Splatting. We also present a new evaluation benchmark for object-centric tracking and reconstruction, offering a complete view of target objects and addressing the limitations of previous benchmarks.

In summary, our contributions are as follows:
\begin{enumerate}
\item An adaptive method for object tracking and reconstruction tailored to varying object geometry and appearance complexity, utilizing 3D Gaussian Splatting. This includes tracking based on both appearance and geometry, as well as a novel key frame selection strategy based on visibility and geodesic distance.
\item A benchmark dataset that provides  full view coverage of objects, along with annotations for challenging scenarios such as yaw movement of axis-symmetrical objects, enabling a comprehensive analysis of state-of-the-art methods in object-centric tracking and reconstruction.
\end{enumerate}

%% file: sec/2_related.tex
\section{Related Works}
In this section, we review prior research on 3D object reconstruction, object pose estimation, and joint tracking and reconstruction.

\subsection{3D Object Reconstruction}
Retrieving a 3D object mesh model from single~\cite{irshad2022shapo, lunayach2023fsd, remelli2020meshsdf, huang2024zeroshape, pix2surf_2020} or multi-view images~\cite{wang2021neus, oechsle2021unisurf, Munkberg_2022_CVPR} has been extensively studied in computer vision. Learning-based single-view shape reconstruction methods rely on categorical priors~\cite{irshad2022shapo, irshad2022centersnap} obtained by pretraining latent shape decoders on a large collection of shapes. Recently, local features enabled zero-shot generalization~\cite{huang2024zeroshape, Iwase_ECCV_2024} to out-of-distribution categories. Other works have utilized recent advances in neural 3D representations and focused on utilizing information from multiple views to enable high-quality 3D reconstructions. While effective, they assume known camera poses. In contrast, our approach focuses on reconstructing high-quality 3D object models from casually captured videos without assuming known camera poses. % \katliu{Missing the generative image-to-3D works?}

\subsection{Object Pose Estimation}
Estimating the orientation and position of objects in a scene is crucial for object manipulation and interaction. Instance-level pose estimation relies on accurate CAD model availability. Works in this domain utilize point~\cite{tekin2018real, rad2017bb8} or dense~\cite{zakharov2019dpod,hodan2020epos, diffusionnocs} correspondences, template matching~\cite{kehl2016deep, sundermeyer2018implicit, tejani2014latent} or directly estimate poses~\cite{kehl2017ssd, wang2019densefusion, xiang2018posecnn}. In contrast, category-level pose estimation does not rely on CAD models during inference, and instead utilizes shape priors~\cite{irshad2022centersnap, lunayach2023fsd, tian2020shape, chen2020learning} or generative models~\cite{diffusionnocs, zhang2024generative}. Recent methods for pose estimation have focused on generalizing to novel objects without fine-tuning. Works in this domain utilize features from foundation models~\cite{caraffa2025freeze, ornek2025foundpose} or via large-scale synthetic training~\cite{foundationposewen2024}. In this work, we focus on object tracking, similar to the motivation of BundleTrack~\cite{wen2021bundletrack} and BundleSDF~\cite{wen2023bundlesdf}. Unlike the above approaches, we jointly track 3D orientation and position of objects while reconstructing high-quality geometry and appearance using 3D Gaussian Splatting~\cite{kerbl20233d}.

\begin{figure*}[t]
\centering
\includegraphics[width=1\linewidth]{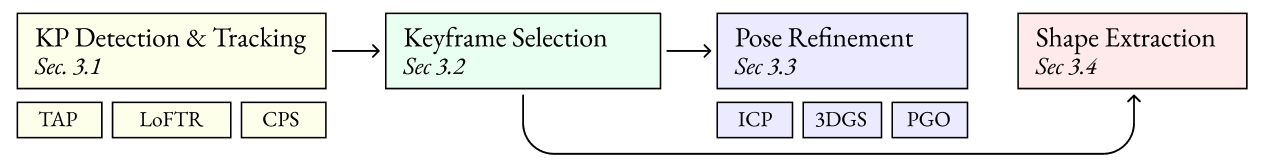}
\caption{~\textbf{Pipeline Flow}: Our method processes sequential RGB-D frames and estimates coarse relative poses using the (1) Keypoint Detection \& Tracking module. If the poses meet the visibility criteria, the frames are added to the keyframe pool (2). To refine the keyframe poses, the (3) Pose Refinement module—incorporating ICP, 3DGS-based render-and-compare, and pose graph optimization techniques—is applied. Finally, shapes are extracted via TSDF fusion using the recovered poses.
}
\label{fig:pipeline}
\end{figure*}

\subsection{Joint Tracking and Reconstruction}
Simultaneous localization and mapping~(SLAM) is a popular area of research~\cite{teed2021droid,murORB2,irshad2024neuralfieldsroboticssurvey} in robotics. Recent works in this domain utilize advances in neural 3D representation and focus on estimating the camera poses while jointly reconstructing the static 3D environment~\cite{Zhu2022CVPR,keetha2024splatam,Matsuki:Murai:etal:CVPR2024}. While effective, they show results on scene-specific settings and assume a static 3D scene. As opposed to scene-level SLAM, object-level SLAM systems~\cite{kong2023vmap} focus on incrementally adding objects online with the recovered poses without assuming any information about the 3D orientation and position of the objects. Dynamic SLAM systems~\cite{henein2020dynamic} explicitly aim to recognize, model, or compensate for dynamic components within a scene. Other works focus on reconstructing individual objects from scene-level incremental posed 2D data either offline or online~\cite{yang2021learningobjectcompositionalneuralradiance,wu2022object,kong2023vmap}. These works utilize object latent codes to learn object-level implicit models. While effective, most approaches utilize complete scene information for tracking and mapping. Our work focuses on joint mapping and tracking of household objects, a promising albeit less studied setting for recovering high-quality 3D object models from casually captured videos. The closest approach to our method is BundleSDF~\cite{wen2023bundlesdf} for object-centric mapping and tracking utilizing a neural object field. BundleSDF struggles with reconstructing the geometry of axis-symmetric objects. In contrast, our approach is able to reconstruct both high-quality object geometry and appearance utilizing a 3D Gaussian Splatting representation.

%% file: sec/3_method.tex
\section{Method}

We aim to track the 6D pose of dynamic unknown objects in a sequence of RGB-D images captured by a single, static camera. The resulting estimated poses are then used for reconstructing a complete textured 3D model. In all our sequences, the objects move within the scene while the camera remains stationary. This setup is more challenging than the more common case where objects are static and the camera moves, as background points in a moving-camera setup provide significantly more information for tracking the scene. 
We assume that three query points, a bounding box, or a mask for the target object are provided in the first frame only. SAM2~\cite{ravi2024sam2} is then used to create the masks for the image sequence.

\begin{figure}[ht]
\centering
\includegraphics[width=1\linewidth]{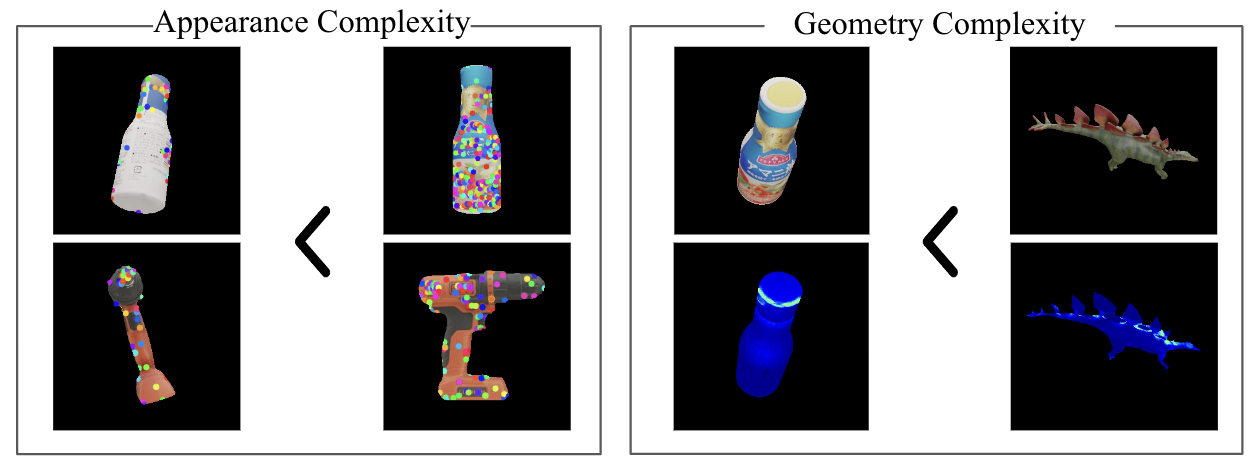}
\caption{~\textbf{Appearance and Geometric Complexity}: The detected keypoints from SIFT are visualized in the left figure. The appearance complexity can vary depending on the viewpoint, even for the same object. Lower complexity makes keypoint tracking more difficult. We also visualize the geometric complexity at each pixel using a color map based on our proposed method. Higher complexity makes the registration problem easier.}
\label{fig:complexity}
\end{figure}

As shown in \cref{fig:pipeline}, the key components of our adaptive algorithm are (1) keypoint detection and tracking, (2) pose refinement, and (3) keyframe selection based on the appearance and geometric complexity as shown in~\cref{fig:complexity}.
Our method utilizes visual cues from both object geometry and appearance, facilitating dynamic adaptation based on the observed data. In the following sections, we provide a detailed discussion of each component.

\subsection{Keypoint Detection and Tracking}
Our method leverages two state-of-the-art off-the-shelf visual trackers: a query-based keypoint tracker, TAP~\cite{doersch2024bootstap}, which is effective for objects with complex appearances, and a detector-free feature matcher, LoFTR~\cite{sun2021loftr}, which performs better in low-texture regions. When a new frame $\mathcal{F}_t$ is introduced, we begin by detecting SIFT features. SIFT (Scale-Invariant Feature Transform) detects keypoints by identifying stable, high-frequency regions across different scales and spatial locations, making it effective for assessing visual complexity. The appearance complexity is thus quantified by evaluating the density of the detected SIFT features. If the feature density falls below a specified threshold (indicating a low-texture region), we use LoFTR~\cite{sun2021loftr} for feature matching. Otherwise, the detected SIFT features are fed into the TAP~\cite{doersch2024bootstap} tracker.

\begin{figure}[b]
\centering
\includegraphics[width=1\linewidth]{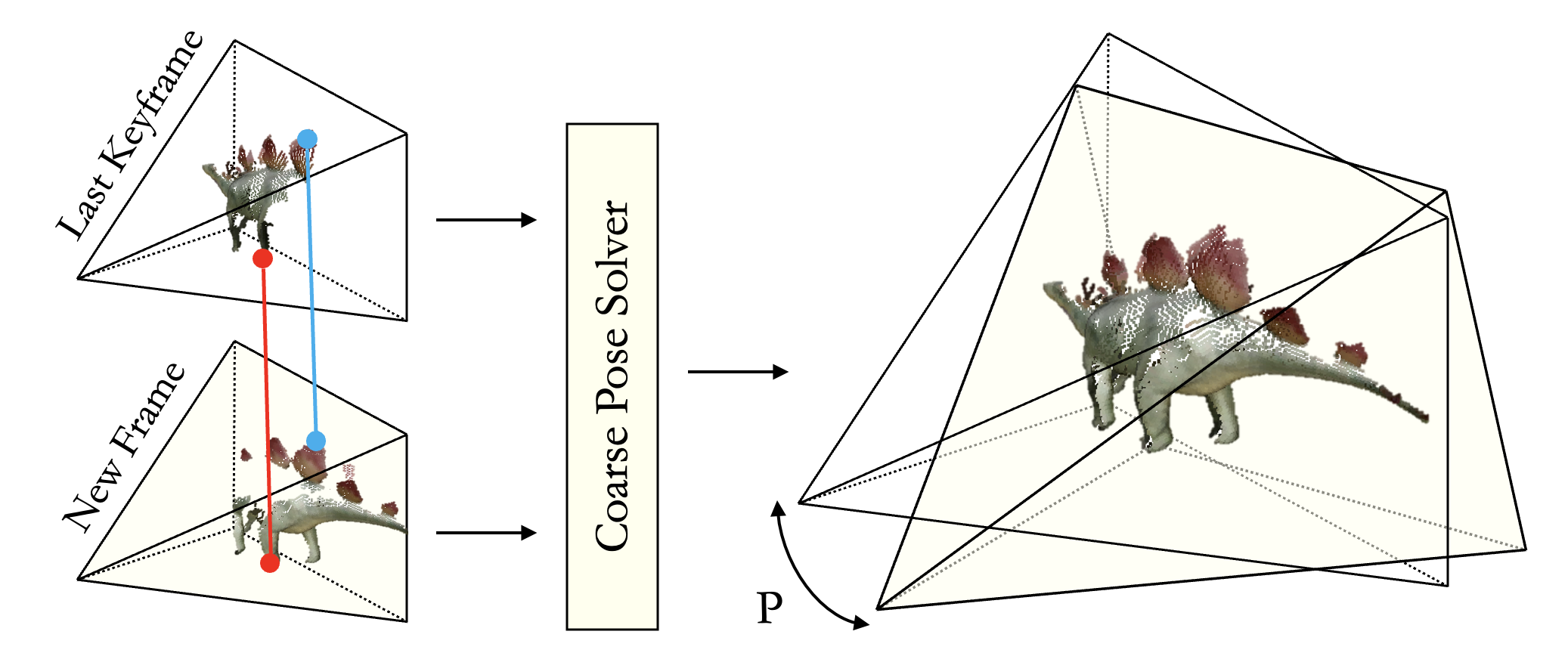}
\caption{~\textbf{Coarse Pose Solver}: 
We use TEASER++ as part of our keypoint detection and tracking module to obtain an initial pose estimate for the new frame relative to the last keyframe using coarse correspondences.}
\label{fig:coarse_pose}
\end{figure}

Then, given correspondences between two consecutive frames, we estimate the relative 6DoF pose $\tilde{\xi}_t$  using a coarse pose solver based on the fast point cloud registration algorithm TEASER++~\cite{teaser} (see \cref{fig:coarse_pose}). This estimated coarse pose is then used for keyframe selection, as described in the next section.

\begin{figure}[t]
\centering
\includegraphics[width=1\linewidth]{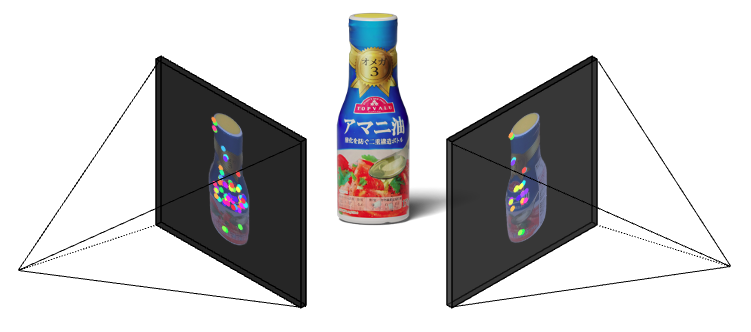}
\caption{~\textbf{Keyframe Selection}: We use the visibility rate across the multiple images based on 2D keypoint tracking as a criteria for keyframe selection. The number of visible key points that is sampled on the left frame decreases when the target object moves.}
\label{fig:kf_selection}
\end{figure}

\subsection{Keyframe Selection}
Simultaneously optimizing the 3D Gaussians, which are used for pose refinement in our pipeline, and camera poses using all images from a video stream is computationally expensive.
So we maintain a keyframe pool $\mathcal{P}$ of selectively chosen frames based on inter-frame co-visibility. Effective keyframe management selects non-redundant frames that observe the same area and span a wide baseline to improve multiview constraints. 

We therefore propose a straightforward keyframe selection method that leverages two factors: (1) the rotational geodesic distance, and (2) the keypoint tracking visibility rate, as shown in \cref{fig:kf_selection}. We begin by storing the first frame $\mathcal{F}_0$ as a keyframe. Next, for each new frame $\mathcal{F}_t$ we iteratively compute these metrics with respect to the last keyframe. If either the rotational distance exceeds a set threshold or the keypoint tracking visibility rate falls below a threshold, $\mathcal{F}_t$ is added to the keyframe pool $\mathcal{P}$. These criteria help reduce redundant information while ensuring sufficient overlap. Additionally, frames with low texture are always included as keyframes due to the increased uncertainty in these cases.

\subsection{Pose Refiner}
If our keyframe selection scheme identifies $\mathcal{F}_t$ as a keyframe, 
we run the keypoint detection and tracking once again between new two keyfames, and proceed to further refine its pose. 
We start with dense point cloud-based registration (ICP~\cite{rusinkiewicz2001efficient}). If the observed geometric complexity is low, we further refine the pose by applying a 3D Gaussian Splatting (3DGS) render-and-compare approach. We skip this step for geometrically complex objects, since, as demonstrated in the experimental section, using 3DGS does not improve pose accuracy and can sometimes introduce additional error in these cases. Additionally, omitting 3DGS on frames with high geometric complexity significantly enhances runtime performance.

\begin{figure}[t]
\centering
\includegraphics[width=1\linewidth]{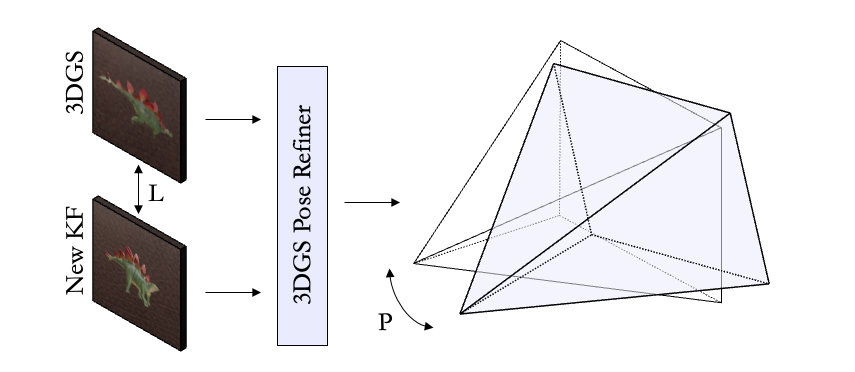}
\caption{~\textbf{3DGS Pose Refiner}: We use 3DGS to improve the initial pose estimate using render-and-compare approach.}
\label{fig:3DGS_pose}
\end{figure}

\paragraph{Geometric Complexity}
To estimate the geometric complexity of the current frame, we compute the mean local curvature of all visible points. The curvature at each point is calculated by first estimating the local covariance matrix \(\mathbf{C}_i\) for each point \(p_i\) based on its surrounding neighbors within a specified radius and maximum number of neighbors. This covariance matrix is then subjected to eigenvalue decomposition, yielding eigenvalues \(\lambda_1, \lambda_2, \lambda_3\), where \(\lambda_1 \leq \lambda_2 \leq \lambda_3\). The local curvature for each point is defined as the ratio of the smallest eigenvalue to the sum of all three eigenvalues, \(\kappa_i = \frac{\lambda_1}{\lambda_1 + \lambda_2 + \lambda_3}\). Finally, the geometric complexity of the frame is represented by the mean of these curvatures across all visible points. 

\begin{figure*}[t]
\centering
\includegraphics[width=1\linewidth]{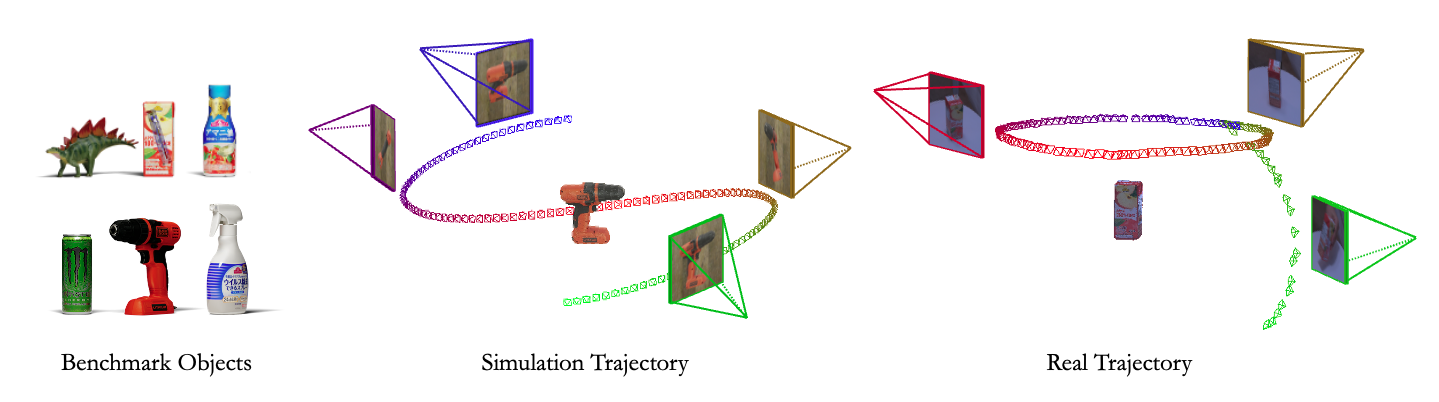}
\caption{~\textbf{GTR3D Benchmark}: Our benchmark includes six objects designed to address some of the most challenging issues faced by existing trackers, such as axis-symmetry, high-frequency geometric details, and diverse materials. To evaluate tracking and reconstruction performance under ideal conditions, we prepare a synthetic dataset (left). To assess performance under real-world conditions, we prepare a real dataset (right) for each object. Both simulated and real trajectories enable full 3D object reconstruction.}
\label{fig:benchmark}
\end{figure*}

\paragraph{3D-GS Pose Refinement}
Coarse key point based pose estimation is insufficient for the high quality tracking when object doesn't contain the rich geometry features, such as axis-symmetrical objects.
In this case we use 3DGS for fine pose refinement by render-and-compare.

To render color $C_i$ at pixel $i$ for each RGB channel, we compute:
\begin{equation}
    C^{ch}_{i} = \sum_{j \leq m}C^{ch}_{j} \cdot \alpha_{j} \cdot T_{j},
\end{equation}
where $C^{ch}_{j}$ represents the color of Gaussian $j$ for the channel $ch$, and $\alpha_j$ and $T_j$ are computed as in depth rendering.  To render depth $D_i$ at pixel $i$ influenced by $m$ ordered Gaussians, we compute:
\begin{equation}
     D_{i} = \sum_{j \leq m}\mu^{z}_{j} \cdot \alpha_{j} \cdot T_{j},
\end{equation}
where $\mu^{z}_{j}$ is the $z$-component of the mean of a 3D Gaussian, with $\alpha_j$ and $T_j$ as weights. We use the same CUDA kernels from~\cite{yugay2023gaussian} for gradient computations over both depth and color, optimizing the parameters of 3D Gaussians while maintaining rendering efficiency. For depth supervision, we define the loss:
\begin{equation}
    L_\text{depth} = |\hat{D} - D|_1,
    \label{eq:depth_loss}
\end{equation}
where $D$ and $\hat{D}$ are the ground-truth and reconstructed depth maps, respectively.

For color supervision, we use a weighted combination of $L_1$ and $\mathrm{SSIM}$~\cite{wang2004image} losses:
\begin{equation}
    L_\text{color} = (1 - \lambda) \cdot |\hat{I} - I|_1 + 
       \lambda \big(1 - \mathrm{SSIM}(\hat{I}, I) \big),
    \label{eq:color_loss}
\end{equation}
where $I$ is the original image, $\hat{I}$ is the rendered image, and $\lambda = 0.2$.

Finally, we combine the color, depth, and regularization terms into a joint loss:
\begin{equation}
    L_{GS} = \lambda_\text{color} \cdot L_\text{color} + \lambda_\text{depth} \cdot L_\text{depth} + \lambda_\text{reg} \cdot L_\text{reg},
\label{eq:joint_loss}
\end{equation}
where $\lambda_\text{color}$, $\lambda_\text{depth}$, and $\lambda_\text{reg}$ are weights for the corresponding losses.

\paragraph{Pose Graph Optimization} \label{sec:pgo}
To reduce the accumulated error, leveraging global consistency is crucial. In our pipeline, sparse point Pose Graph Optimization (PGO) is conducted when a loop closure is detected. Visibility and geodesic distance information are utilized for loop closure detection. First, the geodesic distance between the canonical frame, which is set as the first frame $\mathcal{F}_0$, and the latest frame is calculated. 
Then, the pipeline begins calculating the TAP visibility rate after the geodesic distance exceeds a threshold (set to 180° minus half of our geodesic distance threshold).
If the TAP visibility rate is high, 90\% in our case, or if the geodesic distance is less than 90° while maintaining TAP visibility over 50\%, the PGO is performed. However, PGO is skipped if more than 20\% of the keyframe has low appearance complexity.

In the pose graph \(\mathcal{G} = (\mathcal{V}, \mathcal{E})\), the nodes \(\mathcal{V}\) consist of all the keyframes, \(\mathcal{V} = \{\mathcal{F}_t \mid \mathcal{F}_t \in \mathcal{P}\}\). The edges \(\mathcal{E}\) are defined between neighboring keyframes, such that \(\mathcal{E} = \{ [0, 1], [1, 2], \dots, [N-1, N], [N, 0] \}\). To optimize the poses of all keyframes, the loss function of PGO is defined as follows:

\begin{align} 
\mathcal{L}(i,j)=\sum\limits_{(s_m,s_n) \in E_{i,j}} \rho \left( \left \| \pi_{j}\left(P_{ij} \pi_{i}^{-1} s_{m}\right) - s_{n} \right \|_{2} \right), 
\end{align}

which measures the reprojection error between 2D key point correspondences $s_m,s_n \in E_{i,j}$ between frames $\mathcal{F}_{i}$ and $\mathcal{F}_{j}$ detected using the tracking network, by using the relative transform between the frames $P_{ij}$ estimated using our coarse pose solver. $\pi_j$ denotes the perspective projection mapping onto frame $\mathcal{F}_{j}$, whereas ${\pi^{-1}_{j}}$ represents the inverse projection mapping from frame $\mathcal{F}_{i}$. When the PGO is finished, the canonical frame is updated to be the latest frame. 

\subsection{Shape Extraction} \label{sec:tsdf}
To recover the mesh from the estimated trajectory, we use TSDF fusion. We first fit a 3DGS to all input views using recovered poses. Then we use this 3DGS reconstruction to uniformly render RGB-D views covering the full object using the Icosahedron sampling. Finally, we apply TSDF fusion followed by Marching Cubes to recover the final mesh.

%% file: sec/4_evaluation.tex
\section{Evaluation}

\setlength{\tabcolsep}{10pt}
\begin{table}[t]
  \centering
  \caption{\textbf{Complexity of Benchmark Objects}: Our objects cover a wide range of appearances and geometries, addressing common challenges in tracking and reconstruction.}
    \resizebox{1.0\linewidth}{!}{
    \begin{tabular}{lccc}
    \toprule
         & Axis-Symmetry & Geometry & Texture \\
    \hline
    Box &  &  Low  & High  \\
    Can & \checkmark & Low & Medium \\
    Oil Bottle & \checkmark & Low & High \\
    Spray &  & Medium & Medium \\
    Drill & & Medium & Medium \\
    Dinosaur Toy &  & High & Low \\
    \bottomrule
    \end{tabular}
    }
  \label{tab:obj_complexity}%
\end{table}%

\import{tables/}{sim}

\import{tables/}{real}

\begin{figure*}[t]
\centering
\includegraphics[width=0.8\linewidth]{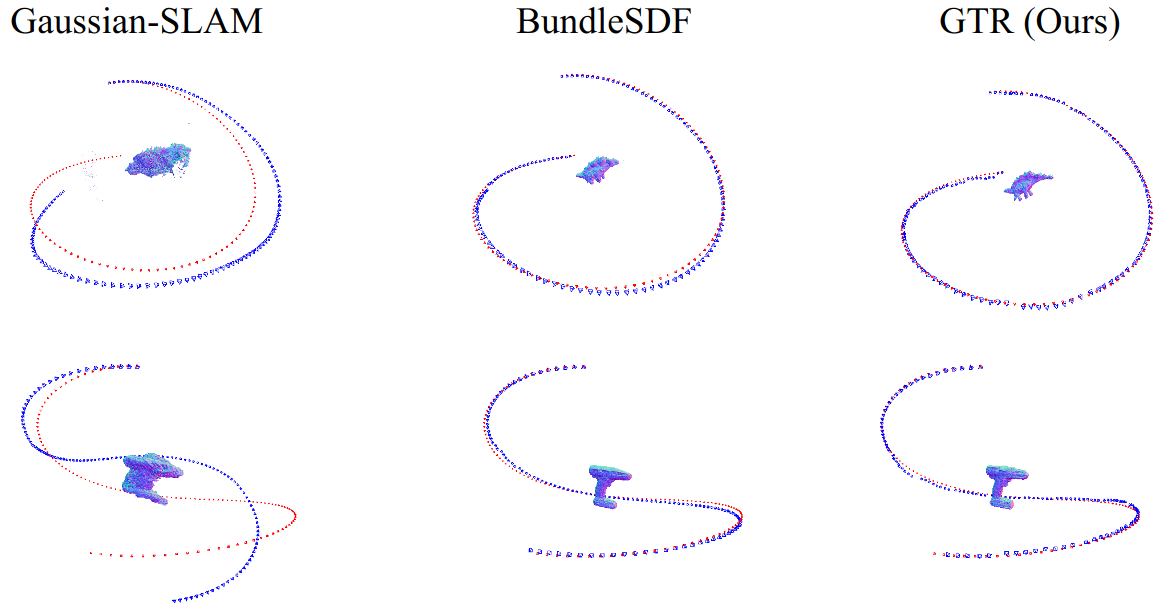}
\caption{~\textbf{Qualitative results on \syndataset}: Our method shows overall best performance across baselines resulting in better object reconstructions. Here red shows ground-truth trajectory and blue shows predicted. We visualize the reconstructed 3D mesh for all approaches.}
\label{fig:synt_qual}
\end{figure*}

\paragraph{GTR3D Benchmark} We evaluate our performance on a newly created benchmark (see~\cref{fig:benchmark}) consisting of six objects: a juice box, oil bottle, energy drink can, drill, spray bottle, and a dinosaur toy. As shown in Tab.~\ref{tab:obj_complexity}, these objects were selected to assess performance from three perspectives: geometric and appearance complexity, material complexity, and axis-symmetry. Tracking and reconstruction quality for each object are evaluated on two sequences: one recorded in simulation with perfect depth and blur-free imagery to assess baseline performance under ideal conditions, and one real sequence recorded by us to evaluate the method's performance in the real world. Synthetic trajectories contain a total of 120 frames, while real trajectories contain 300 frames. We annotate ground truth poses for the real dataset using a combination of FoundationPose~\cite{foundationposewen2024} and ICG+~\cite{icgplus} model-based pose trackers.

\begin{figure*}[t]
\centering
\includegraphics[width=0.8\linewidth]{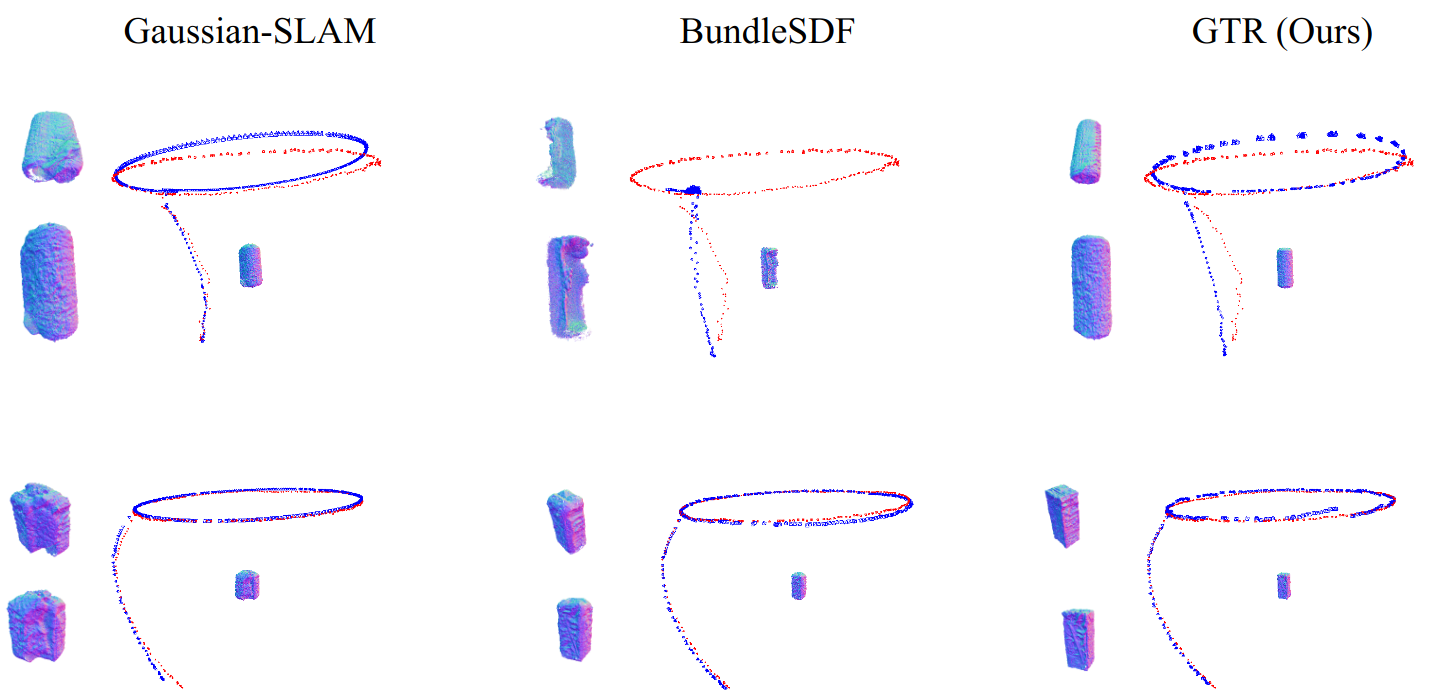}
\caption{~\textbf{Qualitative results on \realdataset}: Our method demonstrates consistent performance for different types of objects resulting in better object reconstructions on average. Here red shows ground-truth trajectory and blue shows predicted. We visualize the reconstructed 3D mesh for all approaches.}
\label{fig:real_qual}
\end{figure*}

\paragraph{Metrics} 
For tracking, we use the average of translation and rotation errors, following \cite{wen2021bundletrack}.
For reconstruction, we use Chamfer Distance (CD) in metric space to evaluate geometry.
To compute CD for all baselines, we use TSDF fusion reconstruction. To do that, we first fit 3DGS to all input views using recovered poses. Then we use this 3DGS reconstruction to uniformly render RGB-D views covering the full object using the Icosahedron sampling using three subdivisions resulting in 162 views per object. Finally, we apply TSDF fusion followed by Marching Cubes to recover the final mesh. 

\paragraph{Baselines}
We consider the following baselines for our method. Gaussian-SLAM~\cite{yugay2023gaussian} is a photo-realistic dense SLAM system based on Gaussian Splatting, enabling high-quality scene representation but without explicit object pose tracking. 
BundleSDF~\cite{wen2023bundlesdf} combines neural 6-DoF tracking with simultaneous 3D geometry refinement of unknown objects, delivering robust tracking in challenging scenarios. Each of these state-of-the-art baselines presents a unique approach to tracking or to combined tracking and reconstruction. We use them to assess the performance and robustness of our method across different aspects of object pose tracking and reconstruction.

We also consider the following baselines: FlowMAP~\cite{smith24flowmap}, Camera-as-Rays~\cite{zhang2024cameras}, and AceZero~\cite{brachmann2025scene}. However, we excluded them due to differences in task settings and incompatibility with our sequential processing setup. Specifically, each of these baselines requires access to the full video sequence from the outset, whereas our approach processes frames sequentially. Additionally, all three baselines operate without depth information or camera intrinsics: Camera-as-Rays relies solely on RGB data, AceZero uses a coordinate prediction network, and FlowMAP finetunes a monocular depth prediction network. To adapt FlowMAP, we provided privileged access to depth information and camera intrinsics; however, the off-the-shelf optical flow model struggled to generalize to our more challenging scenario, where the camera is static, and only the object in the scene moves. Further details are available in the supplementary material.

\subsection{\syndataset}
The goal of the synthetic portion of the benchmark is to evaluate baseline performance in an idealized environment, with perfect depth maps, maximum visibility, and no blur or other camera artifacts.
We show qualitative results in \Cref{tab:synthetic} demonstrating estimated trajectories and object reconstructions.
Our method shows overall best results in both tracking and reconstruction as shown in~\cref{fig:synt_qual}. 
In particular, all our CD scores are the lowest, and only our method demonstrates consistent performance across all objects. 
Although BundleSDF performs well for objects with unique geometries, like the spray can and drill, its tracking scores for axially symmetric objects show high error.

\subsection{\realdataset}
To evaluate baseline performance in a more challenging real-world setting, we use the real portion of our benchmark, which includes sequences of the same six objects used in the synthetic portion. As shown in \Cref{tab:real}, all baselines exhibit inconsistent performance across different object types, whereas our method demonstrates stable performance across all object classes. 
Our method shows the best performance in CD scores, and the translation and rotation errors are within a similar range to GaussianSLAM, which achieves the best average scores for rotation and translation errors. 
Similar to the simulation results, while BundleTrack performs well in translation estimation, its tracking results for rotation are significantly worse. Although BundleSDF excels with objects of unique geometries, such as the spray can and drill, its tracking scores for axially symmetric objects exhibit high errors. On the other hand, GaussianSLAM performs well with axially symmetric objects. However, it shows poor reconstruction performance for geometry-unique objects, such as the dino and spray. We show a qualitative comparison of estimated trajectories and object reconstructions for us and other baselines in~\cref{fig:real_qual}.

\subsection{Ablations}
In \Cref{tab:ablation} we ablate most critical components of our method: 3DGS pose refinement, tracker combinations, and PGO. The full configuration yield the best results in terms of rotation error, while the non-3DGS pose refinement pipeline is the fastest. However, the tracking accuracy of non-3DGS is lower, although the scores for the dino and drill objects remain on par. From a speed perspective, 3DGS refinement can be omitted if necessary. Consequently, the inference time for the dino object is significantly reduced between the full configuration and the setup not evaluating geometric complexity, with improved rotation error.

Additionally, performance differences between TAP and LoFTR trackers are observed for objects like the dino and box. Since the dino has a simpler texture, LoFTR has an advantage, while TAP performs better on well-textured objects like the apple juice box.

In conclusion, iterative pose refinement through 3DGS and global consistency through PGO are crucial for optimal performance, while slowing down inference time.

\import{tables/}{ablation}

%% file: tables/sim.tex
\setlength{\tabcolsep}{20pt}
\begin{table*}
\centering
\resizebox{1\textwidth}{!}{
\begin{tabular}{l|l|rrrrrr||r}
\hline
  \textbf{Method} &
  \multicolumn{1}{c|}{\textbf{Metric}} &
  \textit{box} &
  \textit{dino} &
  \textit{spray} &
  \textit{drill} &
  \textit{can} &
  \textit{oil} &
  Mean
  \\
\hline
  \multirow{3}[2]{*}{Gaussian-SLAM~\cite{yugay2023gaussian}} &
  R\textsubscript{err} $\downarrow$ &
  21.4 &
  21.48 &
  52.96 &
  25.40 &
  38.62 &
  31.01 &
  31.81
  \\
   &
  T\textsubscript{err} $\downarrow$ &
  11.82 &
  26.47 &
  52.69 &
  28.77 &
  24.56 &
  22.04 &
  27.72
  \\
  &
   CD $\downarrow$ &
  11.18 &
  99.2 &
  65.75 &
  33.12 &
  22.64 &
  61.09 &
  48.83
  \\
\hline
\multirow{3}[2]{*}{BundleSDF~\cite{wen2023bundlesdf}} &
  R\textsubscript{err} $\downarrow$ &
  3.23 &
  2.59 &
  \textbf{2.24} &
  \textbf{0.58} &
  90.39 &
  92.38 &
  40.59 \\
   &
  T\textsubscript{err} $\downarrow$ &
  5.27 &
  5.07 &
  \textbf{2.40} &
  \textbf{2.13} &
  \textbf{4.05} &
  2.57 &
  \textbf{3.33} \\
  &
   CD $\downarrow$ &
  4.03 &
  3.59 &
  1.99 &
  2.29 &
  5.86 &
  6.43 &
  4.03
  \\
\hline
  \multirow{3}[2]{*}{Ours} &
  R\textsubscript{err} $\downarrow$ &
  \textbf{1.10} &
  \textbf{1.50} &
  9.35 &
  3.49 &
  \textbf{33.43} &
  \textbf{1.94} &
  \textbf{8.47} \\
   &
  T\textsubscript{err} $\downarrow$ &
  \textbf{0.47} &
  \textbf{1.02} &
  7.54 &
  3.21 &
  18.95 &
  \textbf{1.19} &
  5.40\\
  &
   CD $\downarrow$ &
  \textbf{0.50} &
  \textbf{1.60} &
  \textbf{1.91} &
  \textbf{1.35} &
  \textbf{3.13} &
  \textbf{2.54} &
  \textbf{1.83}
  \\
\bottomrule
\end{tabular}%
}
\caption{\textbf{Results on \syndataset}: For the metrics of R\textsubscript{err}, T\textsubscript{err}, and CD lower value is better.} % For the metrics PSNR, higher is better.}
\label{tab:synthetic}
\end{table*}

%% file: tables/real.tex
\setlength{\tabcolsep}{20pt}
\begin{table*}
\centering
\resizebox{1\textwidth}{!}{
\begin{tabular}{l|l|rrrrrr||r}
\hline
  \textbf{Method} &
  \multicolumn{1}{c|}{\textbf{Metric}} &
  \textit{box} &
  \textit{dino} &
  \textit{spray} &
  \textit{drill} &
  \textit{can} &
  \textit{oil} &
  Mean
  \\
\hline
  \multirow{3}[2]{*}{Gaussian-SLAM~\cite{yugay2023gaussian}} &
  R\textsubscript{err} $\downarrow$ &
  3.54 &
  10.2 &
  5.34 &
  4.16 &
  \textbf{3.95} &
  6.51 &
  \textbf{5.61}
  \\
   &
  T\textsubscript{err} $\downarrow$ &
  3.38 &
  13.7 &
  6.88 &
  \textbf{3.94} &
  \textbf{3.73} &
  5.26 &
  6.15
  \\
  &
   CD $\downarrow$ &
  12.68 &
  44.8 &
  20.27 &
  9.89 &
  11.90 &
  13.57 &
  18.85
  \\
\hline
  \multirow{3}[2]{*}{BundleSDF~\cite{wen2023bundlesdf}} &
  R\textsubscript{err} $\downarrow$ &
  2.98 &
  \textbf{1.55} &
  \textbf{3.98} &
  \textbf{3.53} &
  67.21 &
  67.09 &
  24.39 \\
   &
  T\textsubscript{err} $\downarrow$ &
  \textbf{2.22} &
  7.75 &
  \textbf{2.70} &
  7.17 &
  4.17 &
  \textbf{1.66} &
  \textbf{4.28} \\
  &
   CD $\downarrow$ &
  3.47 &
  33.5 &
  \textbf{10.3} &
  7.66 &
  5.49 &
  35.74 &
  16.02
  \\
\hline
  \multirow{3}[2]{*}{Ours} &
  R\textsubscript{err} $\downarrow$ &
  \textbf{2.47} &
  8.97 &
  12.10 &
  6.51 &
  11.37 &
  \textbf{4.76} &
  7.70 \\
 &
  T\textsubscript{err} $\downarrow$ &
  2.35 &
  \textbf{6.61} &
  9.78 &
  5.66 &
  9.44 &
  4.45 &
  6.37 \\
  &
   CD $\downarrow$ &
  \textbf{2.34} &
  \textbf{8.00} &
  45.20 &
  \textbf{7.47} &
  \textbf{5.24} &
  \textbf{3.80} &
  \textbf{12.01}
  \\
\bottomrule
\end{tabular}%
}
\caption{\textbf{Results on \realdataset}: For the metrics of R\textsubscript{err}, T\textsubscript{err}, and CD lower value is better.}
\label{tab:real}
\end{table*}

%% file: tables/ablation.tex
\setlength{\tabcolsep}{6pt}
\begin{table}[t]
\centering
\resizebox{1\linewidth}{!}{
\begin{tabular}{l|r|rrrrrr||r}
\hline
  \textbf{Method} &
  \textbf{Metric} &
  \textit{box} &
  \textit{dino} &
  \textit{spray} &
  \textit{drill} &
  \textit{can} &
  \textit{oil} &
  Mean
  \\
\hline
 \multirow{2}[2]{*}{No 3DGS} &
  R\textsubscript{err} $\downarrow$ &
  6.04 &
  1.51 &
  19.15 &
  \textbf{3.12} &
  58.83 &
  33.59 &
  20.37 \\
  &
  Time $\downarrow$ &
  \textbf{1.75} &
  \textbf{1.12} &
  \textbf{1.21} &
  \textbf{1.19} &
  \textbf{3.26} &
  \textbf{4.28} &
  \textbf{2.17} \\
\hdashline
 \multirow{2}[2]{*}{No App. Complexity [LoFTR]} &
  R\textsubscript{err} $\downarrow$ &
  4.37 &
  \textbf{0.94} &
  18.91 &
  6.35 &
  \textbf{21.23} &
  2.37 &
  9.03 \\
  &
  Time $\downarrow$ &
  78.60 &
  5.32 &
  78.37 &
  12.85 &
  97.46 &
  131.7 &
  67.39 \\
\hdashline
 \multirow{2}[2]{*}{No App. Complexity [TAP]} &
  R\textsubscript{err} $\downarrow$ &
  1.04 &
  1.99 &
  10.84 &
  3.53 &
  32.16 &
  1.94 &
  8.59 \\
  &
  Time $\downarrow$ &
  15.57 &
  3.17 &
  14.48 &
  4.47 &
  73.23 &
  291.7 &
  67.12 \\
\hdashline
 \multirow{2}[2]{*}{No Geo. Complexity} &
  R\textsubscript{err} $\downarrow$ &
  \textbf{0.95} &
  6.96 &
  9.36 &
  3.32 &
  33.09 &
  \textbf{1.79} &
  9.24 \\
  &
  Time $\downarrow$ &
  16.38 &
  34.41 &
  18.08 &
  17.37 &
  73.00 &
  % 292.79 &
  292.8 &
  75.34 \\
\hdashline
 \multirow{2}[2]{*}{No PGO} &
  R\textsubscript{err} $\downarrow$ &
  1.12 &
  1.43 &
  \textbf{9.21} &
  3.42 &
  33.31 &
  11.86 &
  10.06 \\
  &
  Time $\downarrow$ &
  6.04 &
  1.51 &
  19.15 &
  3.12 &
  58.83 &
  33.59 &
  20.37 \\
\hline
 \multirow{2}[2]{*}{Full Configuration} &
  R\textsubscript{err} $\downarrow$ &
  1.10 &
  1.50 &
  9.35 &
  3.49 &
  33.43 &
  1.94 &
  \textbf{8.47} \\
  &
  Time $\downarrow$ &
  15.08 &
  3.56 &
  16.83 &
  4.60 &
  73.07 &
  286.3 &
  66.58 \\
\bottomrule
\end{tabular}%
}
\caption{\textbf{Ablation Study on \syndataset}: For the metrics of Rerr, Time [min], lower value is better.}
\label{tab:ablation}
\end{table}

%% file: sec/5_conclusion.tex
\section{Conclusion}
We present a method that advances 6-DoF object tracking and high-quality 3D reconstruction from monocular RGB-D video, addressing limitations of existing approaches and providing a stable solution for handling objects with diverse properties—including symmetry, intricate appearances and high-frequency geometry—whereas other methods often prioritize specific object classes. By integrating 3D Gaussian Splatting, hybrid geometry/appearance tracking, and keyframe selection, we achieve robust tracking and detailed reconstructions across a wide range of challenging objects. To support further research, we also introduce a benchmark with high-quality annotations for evaluating both tracking and reconstruction performance on these difficult object classes.

%% file: sec/X_suppl.tex
\clearpage
\setcounter{page}{1}
\maketitlesupplementary

\appendix

\section{Details of GTR3D Real Dataset}
\label{sec:real_dataset}
\begin{figure}[t]
\centering
\includegraphics[width=1\linewidth]{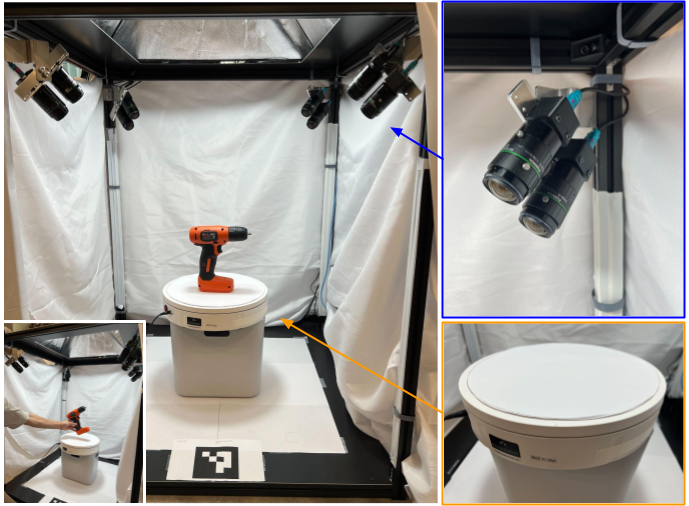}
\caption{Equipment for GTR3D Real Creation 
}
\label{fig:scan_box}
\end{figure}

% \the\columnwidth
There is no public dataset containing an accurate series of RGBD images that covers a large variety of objects with full views. These features are crucial for investigating and evaluating the performance of object-centric tracking and reconstruction.
As such, we created a custom real dataset using the equipment shown in \cref{fig:scan_box}. To capture full views of the target objects, they were manipulated using a turntable and human hand. In total, a series of 300 images was captured using a single set of stereo cameras.

\begin{figure}[thb]
\centering
\includegraphics[width=1\linewidth]{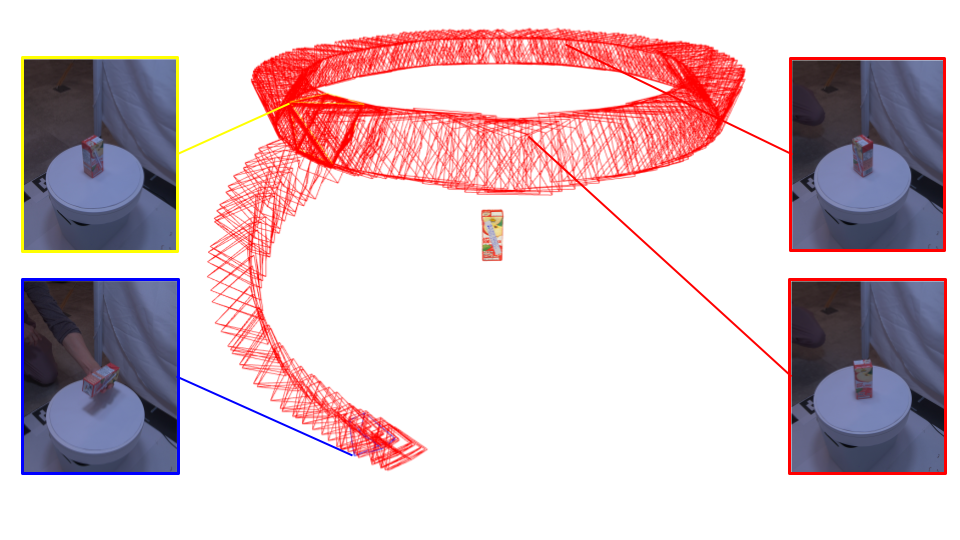}
\caption{Object Trajectory in GTR3D Real 
}
\label{fig:real_traj}
\end{figure}

\begin{figure*}[!t]
\centering
\includegraphics[width=0.9\linewidth]{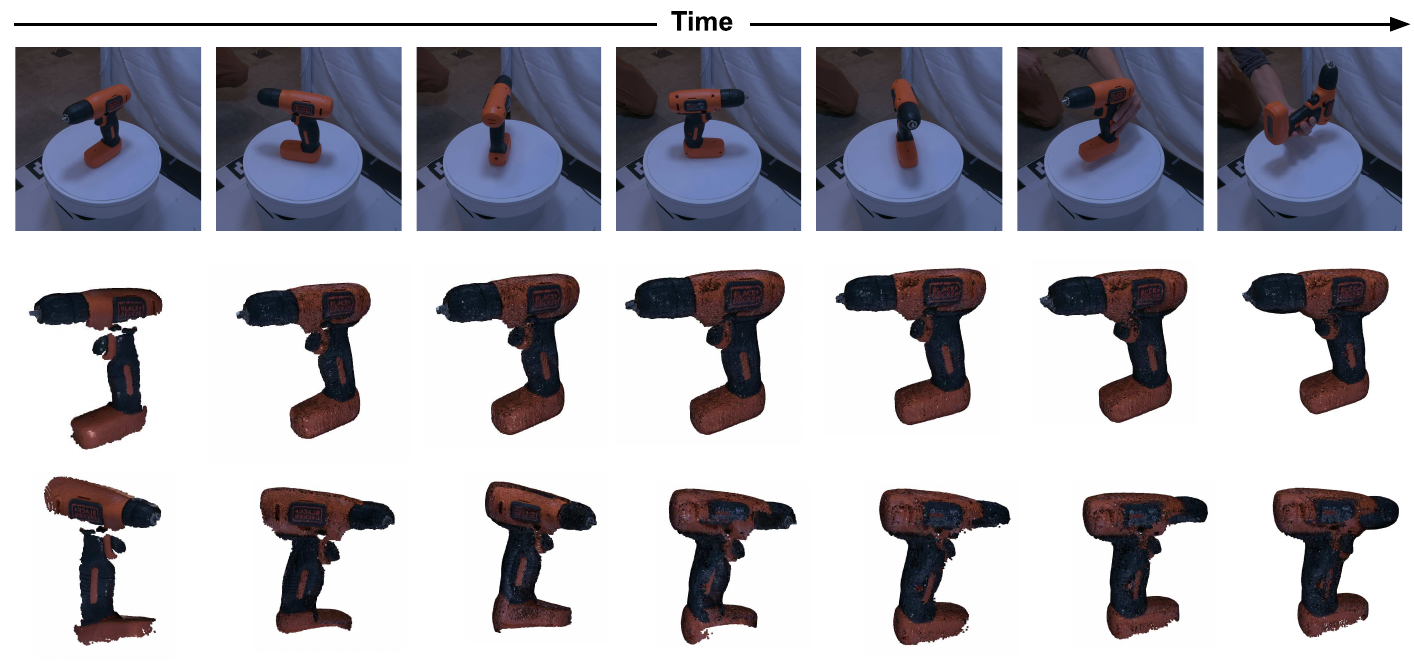}
\caption{Tracking and Reconstruction on GTR3D Real
}
\label{fig:sup_teaser}
\end{figure*}

In ~\cref{fig:real_traj}, the trajectory from the object's coordinates is visualized. The image outlined in yellow represents the first frame. After the object completes a 360-degree rotation on the turntable, it is manipulated by hand to expose the bottom of the object to the camera. The image outlined in blue represents the last frame. ~\cref{fig:sup_teaser} also presents a series of images captured sequentially over time. For more details about the object's motion, refer to the supplementary video.

For data collection, we used a stereo camera with a global shutter and a resolution of 24 MP. The original resolution of 4608 × 5328 was rectified and downscaled to 2048 × 2560. The learned stereo method \cite{shankar2022learned} was then employed to generate depth images from the stereo images.

In total, 300 images were captured at 7.5 fps over a 40-second sequence. The turntable completes a full rotation in 30 seconds, and the bottom of the target object is shown to the camera by hand in the remaining 10 seconds.

\section{Pose Annotations of GTR3D Real Dataset}
\label{sec:real_dataset}

\begin{figure}[b]
\centering
\includegraphics[width=1\linewidth]{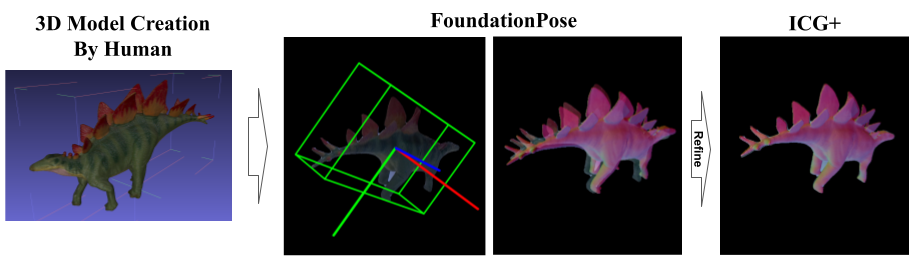}
\caption{6D Pose Annotation
}
\label{fig:annotation}
\end{figure}

To evaluate tracking performance, pose annotations are required for the real data. For this, we first created 3D models with the help of digital artists. Using the created 3D models, coarse poses were estimated using FoundationPose \cite{foundationposewen2024}. These poses were then refined with ICG+ \cite{icgplus}, which utilizes data from four stereo camera setups to improve tracking performance by combining information from multiple cameras, as shown in \cref{fig:scan_box} and \cref{fig:annotation}.

However, we observed that the performance of ICG+ in tracking yaw movement (caused by the turntable) for axially symmetric objects was not stable. Consequently, we retained the results from FoundationPose for the axially symmetric objects: the oil bottle and the energy drink can.

% \begin{figure*}[!t]
% \centering
% \includegraphics[width=0.9\linewidth]{fig/qualitative_recon_1.pdf}
% \caption{~\textbf{Reconstruction results}
% }
% \label{fig:recon_1}
% \end{figure*}

% \begin{figure*}[!t]
% \centering
% \includegraphics[width=0.9\linewidth]{fig/qualitative_recon_2.pdf}
% \caption{~\textbf{Reconstruction results}
% }
% \label{fig:recon_1}
% \end{figure*}

% \begin{figure*}[!t]
% \centering
% \includegraphics[width=0.9\linewidth]{fig/qualitative_recon_3.pdf}
% \caption{~\textbf{Reconstruction results}
% }
% \label{fig:recon_1}
% \end{figure*}

\section{Online and Offline Processes}
\label{sec:onlie_offline}

As part of the online process, keypoint detection and tracking, along with pose refinement, are performed iteratively. For the 3DGS render-and-compare step, a single 3DGS is iteratively reconstructed from each RGB-D image and used to perform render-and-compare for pose refinement between the previous and latest keyframes.

Pose graph optimization, however, is not performed iteratively; it is executed only when the conditions described in Sec. 3.3 of the online process are satisfied.

In the offline process, the full 3DGS is reconstructed using all keyframes. This 3DGS is then utilized to create the mesh model through TSDF fusion, as described in Sec. 3.4.

\begin{figure}[b]
\centering
\includegraphics[width=1\linewidth]{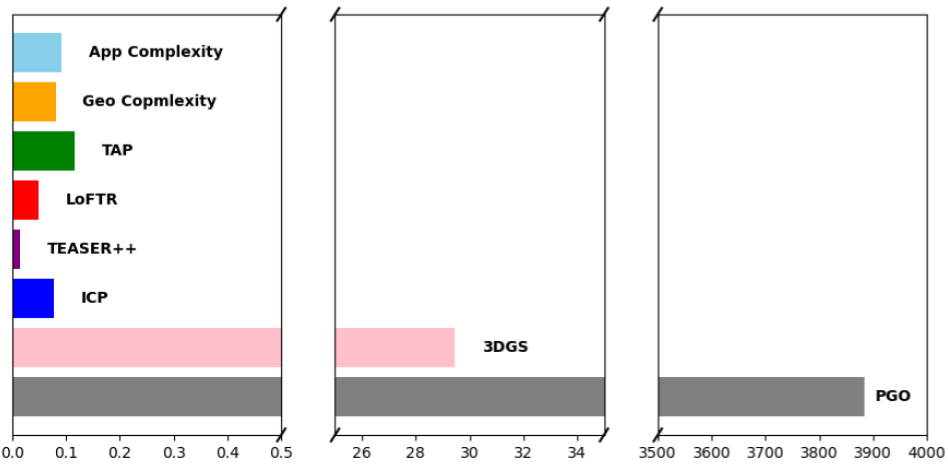}
\caption{Execution Time Comparison
}
\label{fig:inf_time}
\end{figure}

\section{Execution time of online process} \label{inf_time}

In ~\cref{fig:inf_time}, the execution times of the core components for the oil bottle using the synthetic dataset are visualized: calculation of appearance complexity (0.092s), calculation of geometry complexity (0.081s), TAP \cite{doersch2024bootstap} (0.116s), LoFTR \cite{sun2021loftr} (0.048s), TEASER++ \cite{teaser} (0.014s), ICP (0.077s), 3DGS reconstruction and render-and-compare (29.471s), and PGO (7959.463s). Among these, 3DGS reconstruction and PGO are the two most time-consuming components. For the non-linear optimization of PGO, we employed a simple implementation using the PyPose library \cite{wang2023pypose}.

\section{Reconstruction Results}
\label{sec:qual_recon_eval}

We show our reconstruction results on GTR3D Real from both appearance and geometry perspectives, comparing them with 2 SOTA baselines: 
% BundleTrack \cite{wen2021bundletrack}, 
BundleSDF \cite{wen2023bundlesdf}, Gaussian-SLAM \cite{yugay2023gaussian}. 

\begin{figure*}[tb]
\centering
\includegraphics[width=1\linewidth]{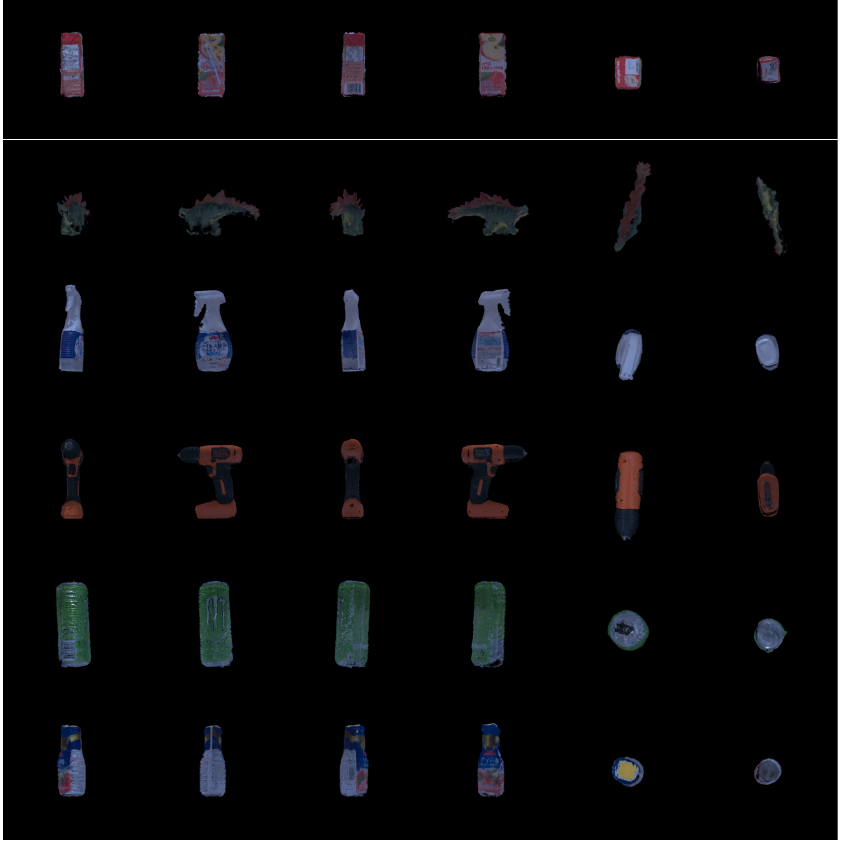}
\caption{Our 3DGS Reconstructions on GTR3D Real
}
\label{fig:all_app_recon}
\end{figure*}

\begin{figure*}[!t]
\centering
\includegraphics[width=0.9\linewidth]{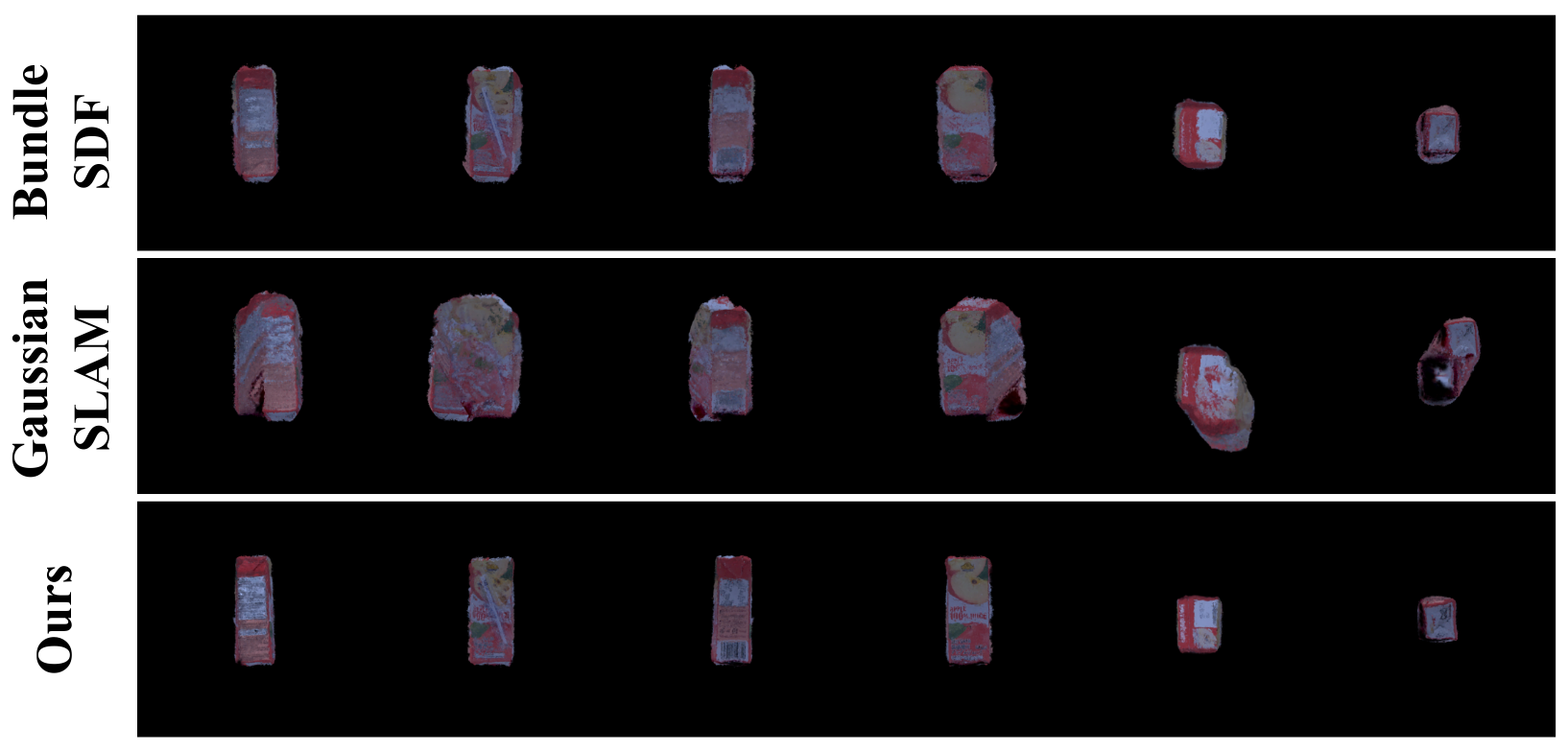}
\caption{3DGS Reconstruction Comparison on GTR3D Real, Box Object
}
\label{fig:box_app_recon}
\end{figure*}

\begin{figure*}[!b]
\centering
\includegraphics[width=0.9\linewidth]{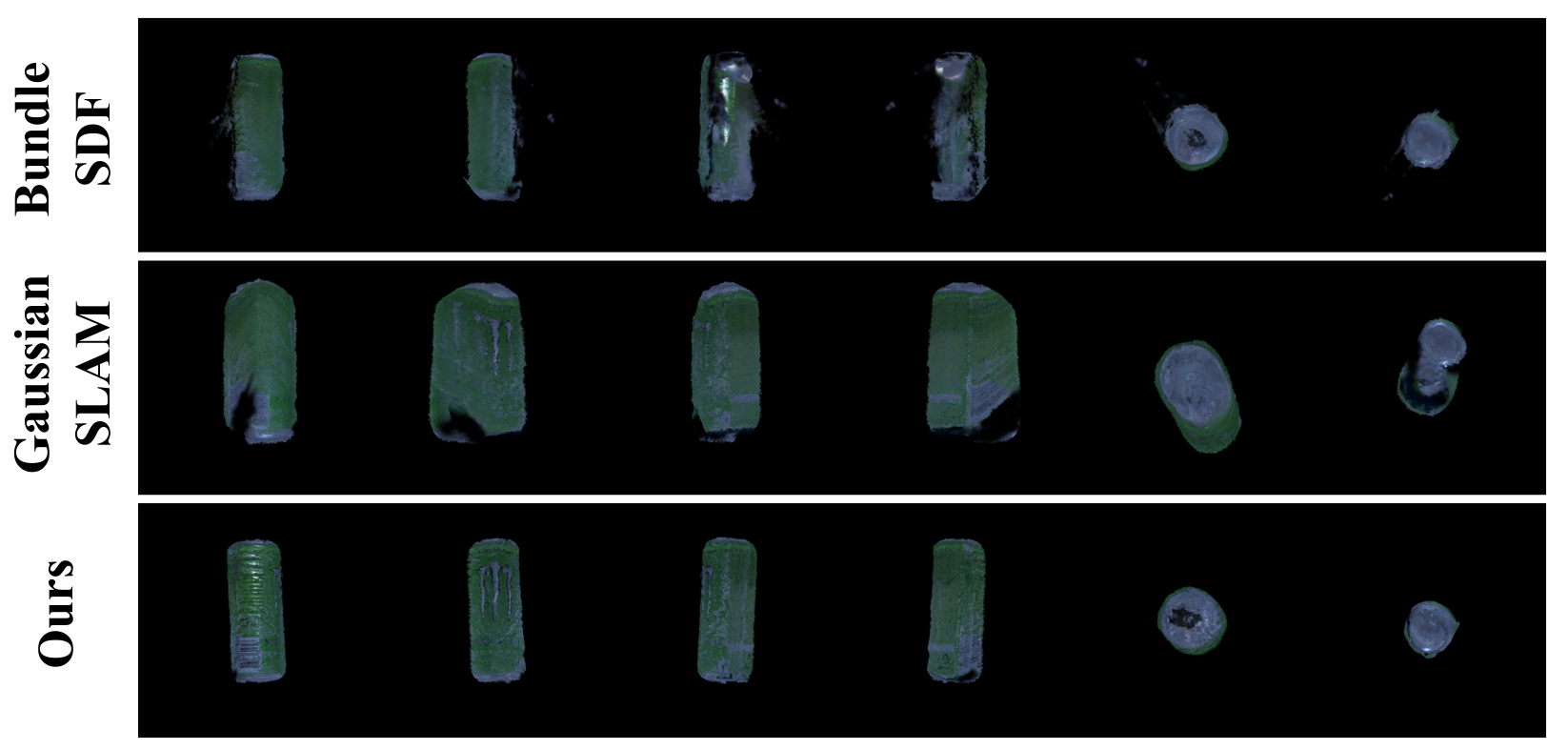}
\caption{3DGS Reconstruction Comparison on GTR3D Real, Can Object
}
\label{fig:can_app_recon}
\end{figure*}

\begin{figure*}[!t]
\centering
\includegraphics[width=0.9\linewidth]{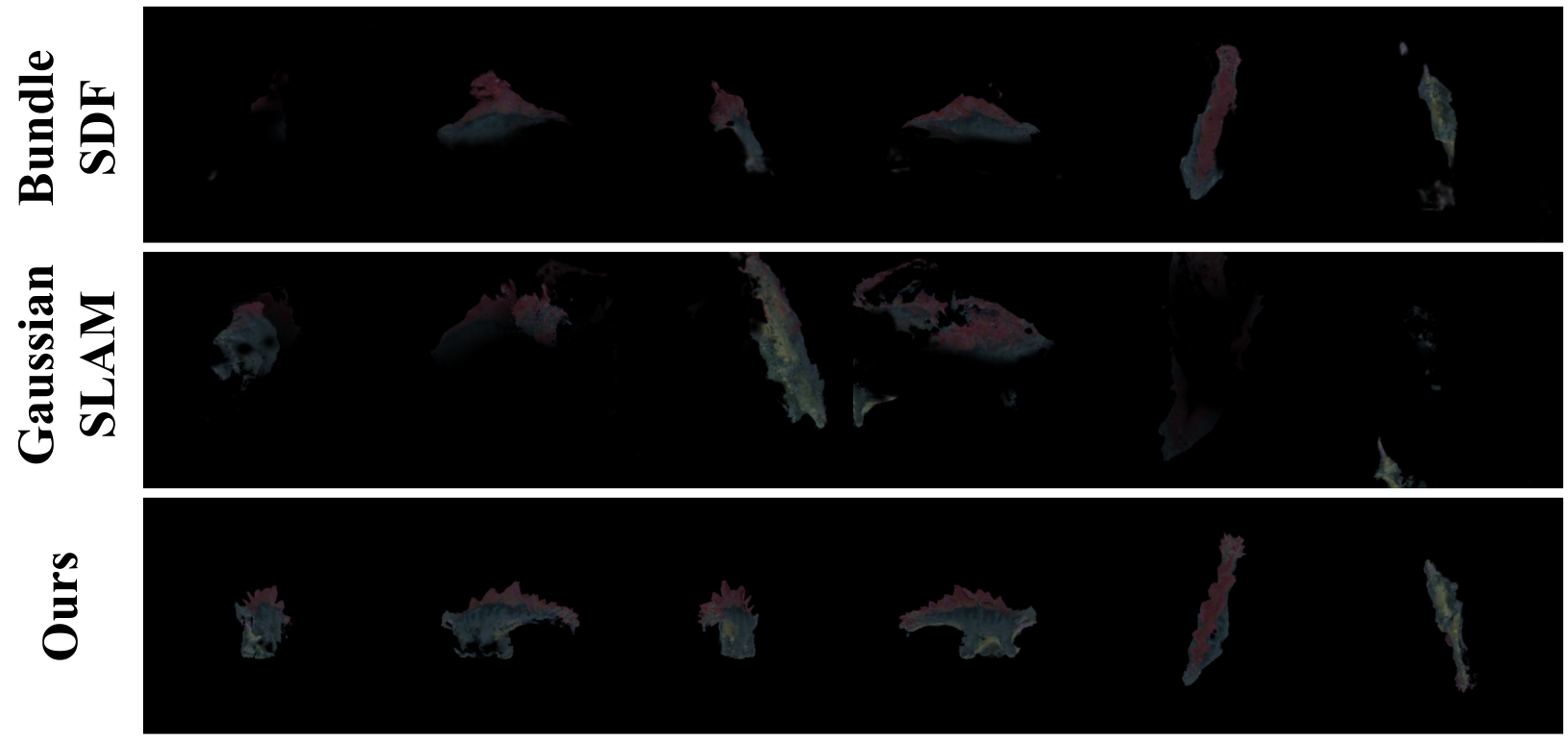}
\caption{3DGS Reconstruction Comparison on GTR3D Real, Dino Object
}
\label{fig:dino_app_recon}
\end{figure*}

\begin{figure*}[!b]
\centering
\includegraphics[width=0.9\linewidth]{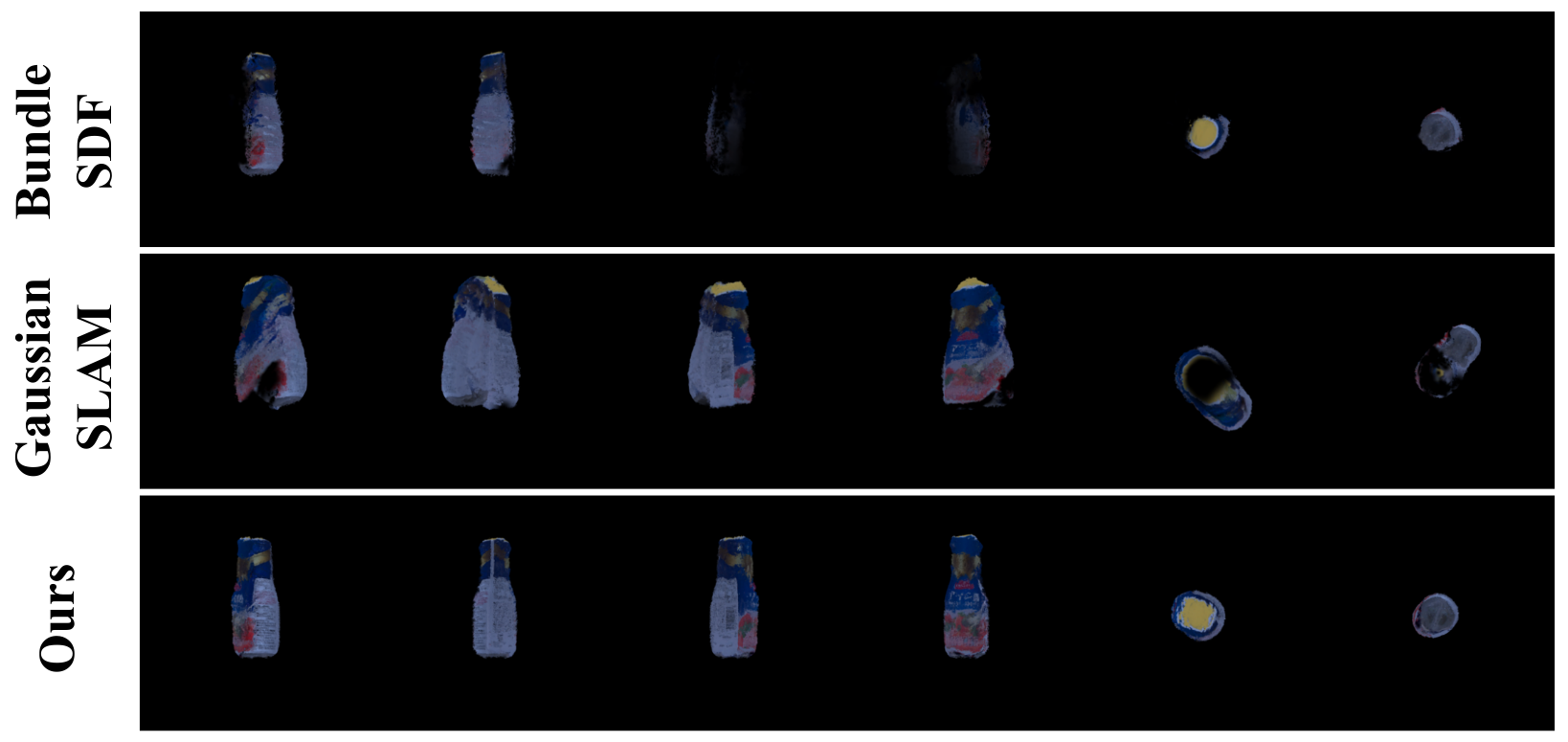}
\caption{3DGS Reconstruction Comparison on GTR3D Real, Bottle Object
}
\label{fig:oil_app_recon}
\end{figure*}

\begin{figure*}[!t]
\centering
\includegraphics[width=0.9\linewidth]{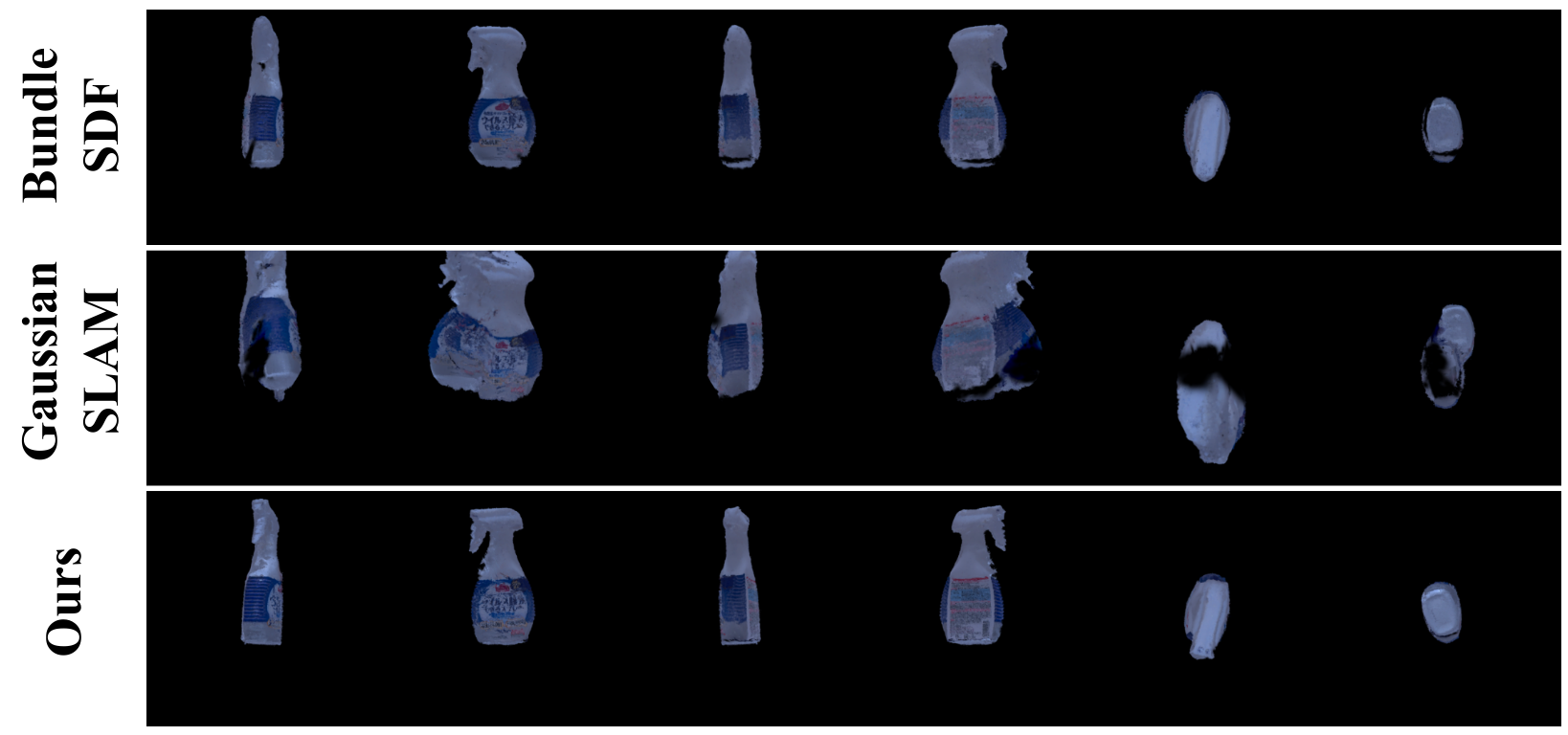}
\caption{3DGS Reconstruction Comparison on GTR3D Real, Spray Object
}
\label{fig:spray_app_recon}
\end{figure*}

\begin{figure*}[!b]
\centering
\includegraphics[width=0.9\linewidth]{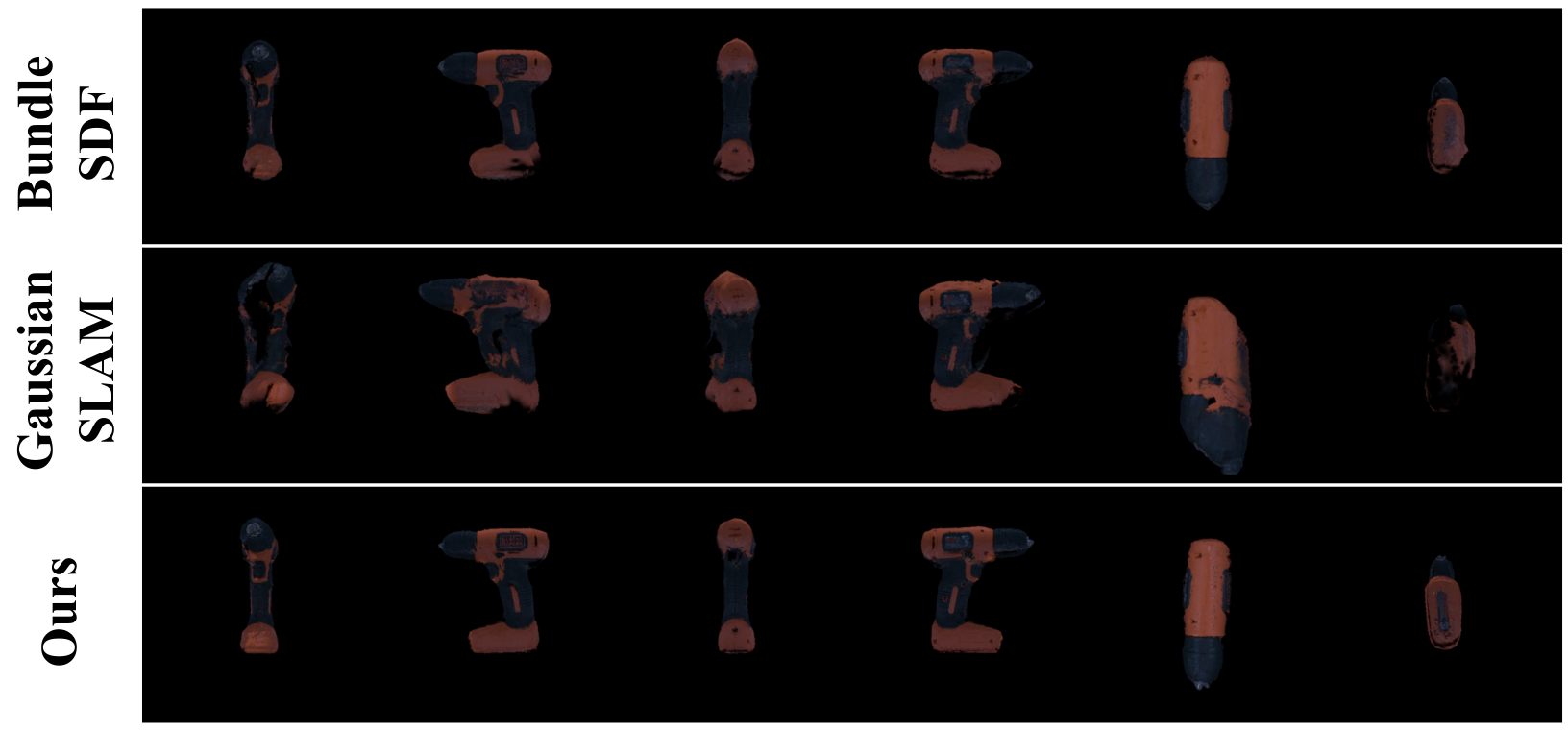}
\caption{3DGS Reconstruction Comparison on GTR3D Real, Drill Object
}
\label{fig:drill_app_recon}
\end{figure*}

In the \cref{fig:all_app_recon}, we render images from 6 orthogonal views (left, front, right, back, top, bottom) for all 6 objects via 3DGS. To further compare the quality between our method and baselines, we tiled rendered images for all 6 objects in \cref{fig:box_app_recon}, \cref{fig:can_app_recon}, \cref{fig:dino_app_recon}, \cref{fig:oil_app_recon}, \cref{fig:spray_app_recon}, \cref{fig:drill_app_recon}. 
The quality of GaussianSLAM is unstable for all objects. 
% For BundleTrack and 
For BundleSDF, quality for axis-symmetric objects, such as the can and bottle, is lower than non symmetric objects because partial areas (right and back) are not well reconstructed due to failed tracking. On the other hand, our method shows stable performance for all objects.

\begin{figure*}[!t]
\centering
\includegraphics[width=0.8\linewidth]{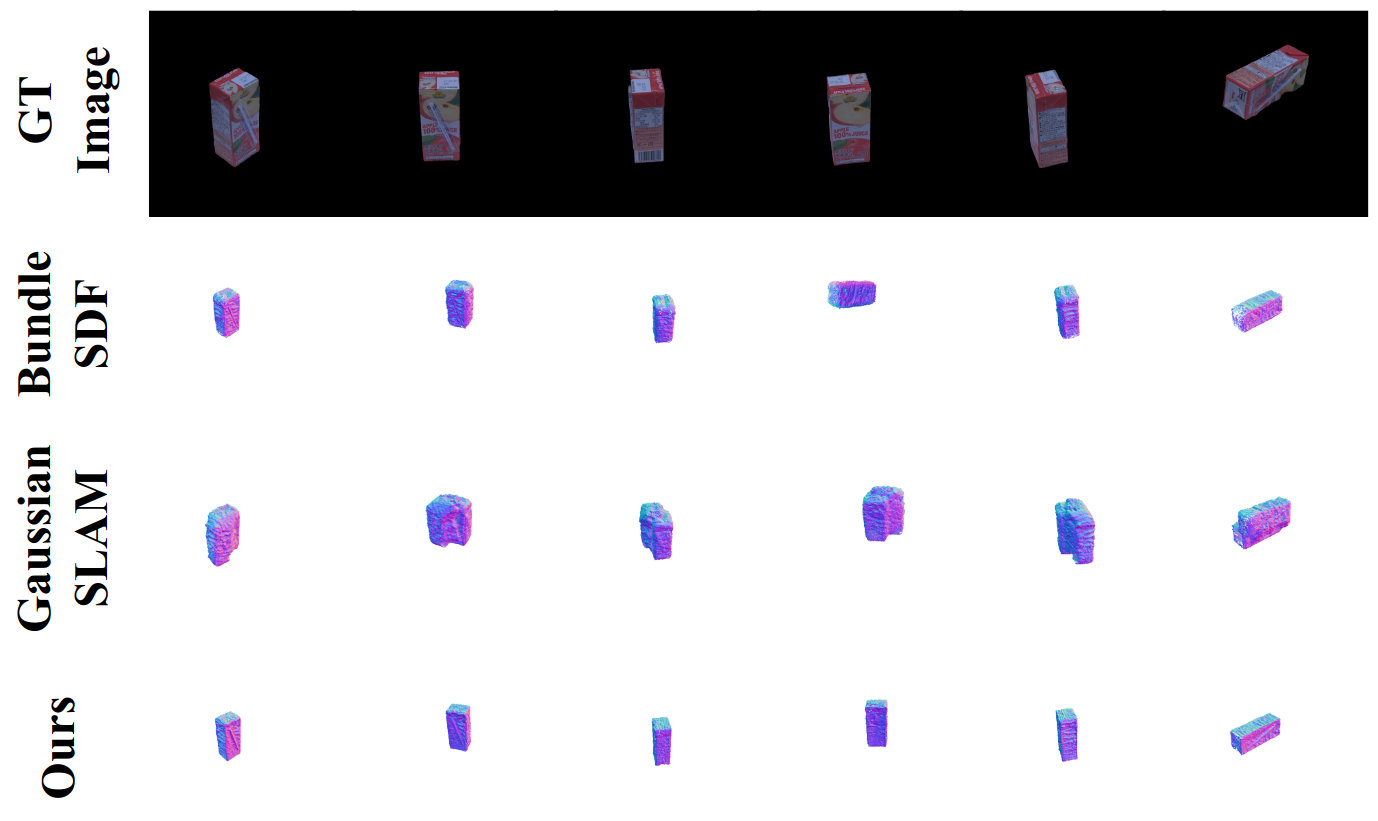}
\caption{Mesh Reconstruction Comparison on GTR3D Real, Box Object 
}
\label{fig:box_tsdf}
\end{figure*}

\begin{figure*}[!b]
\centering
\includegraphics[width=0.8\linewidth]{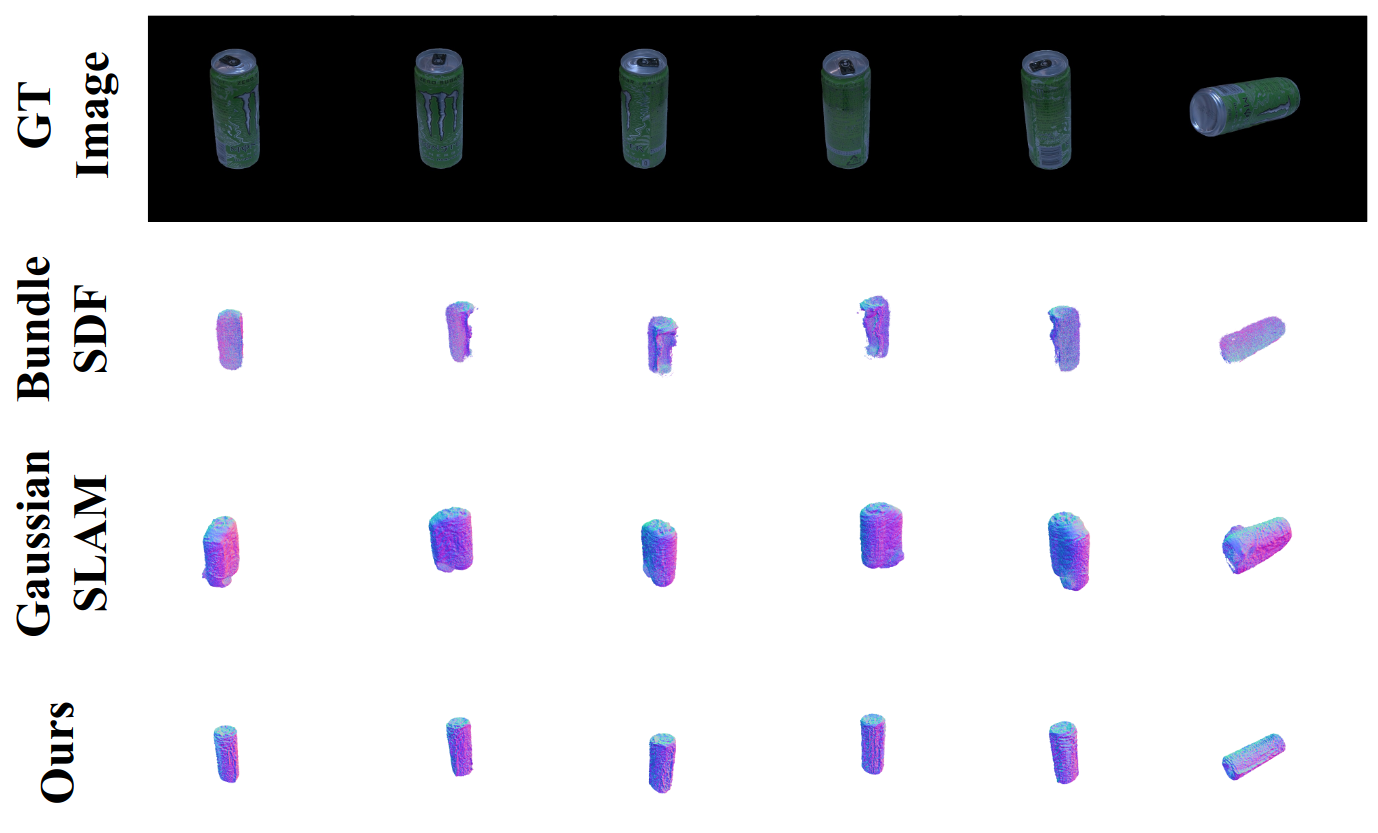}
\caption{Mesh Reconstruction Comparison on GTR3D Real, Can Object 
}
\label{fig:can_tsdf}
\end{figure*}

\begin{figure*}[!t]
\centering
\includegraphics[width=0.8\linewidth]{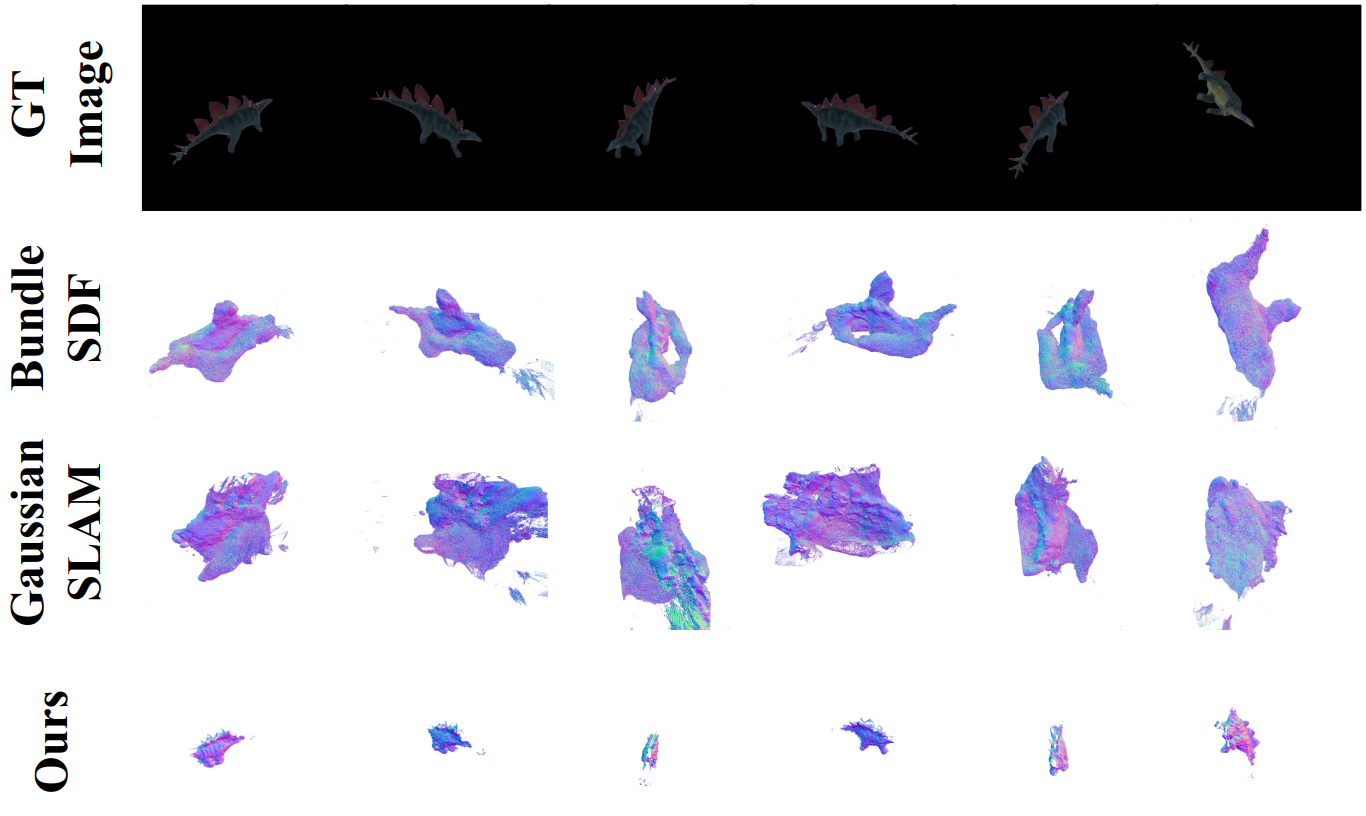}
\caption{Mesh Reconstruction Comparison on GTR3D Real, Dino Object 
}
\label{fig:dino_tsdf}
\end{figure*}

\begin{figure*}[!b]
\centering
\includegraphics[width=0.8\linewidth]{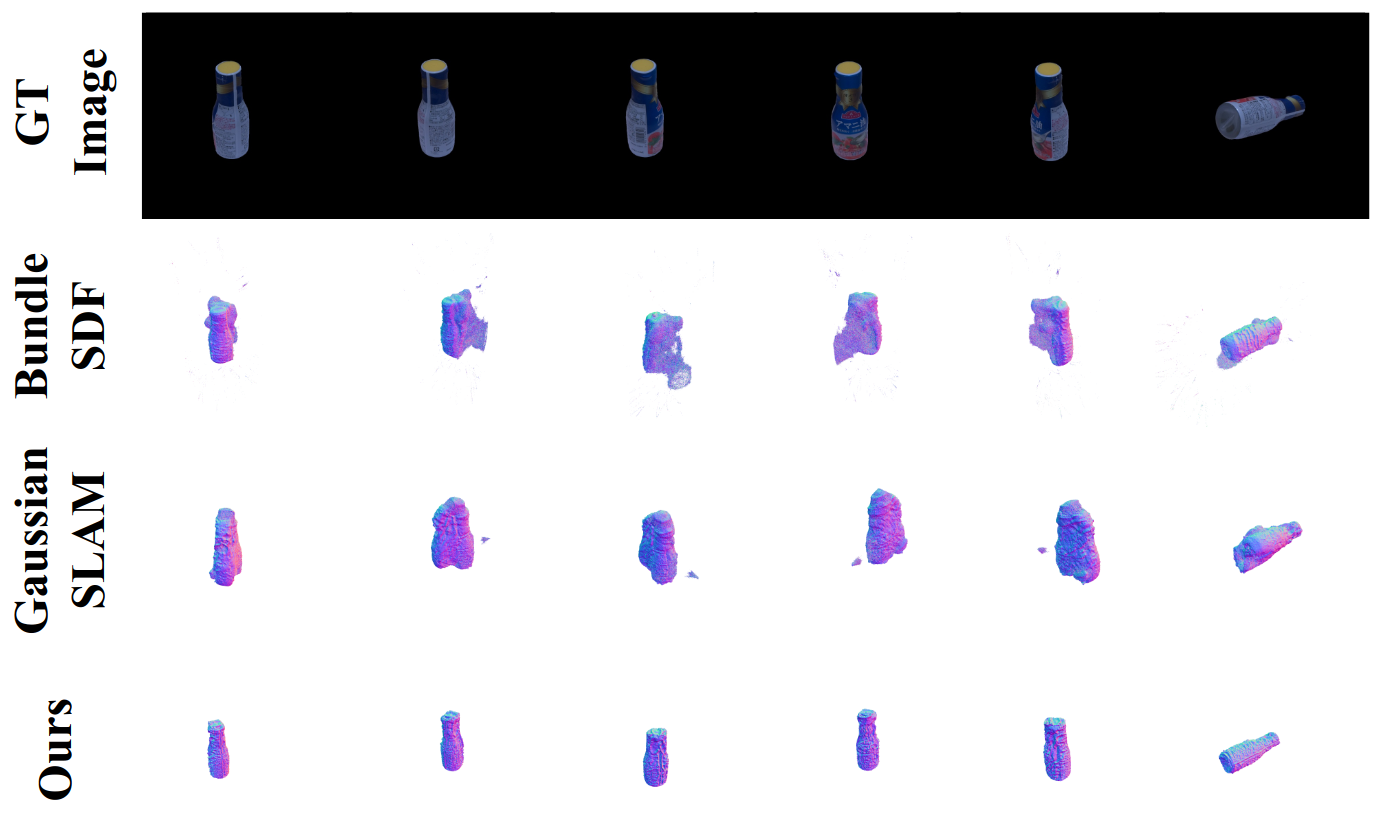}
\caption{Mesh Reconstruction Comparison on GTR3D Real, Bottle Object 
}
\label{fig:oil_tsdf}
\end{figure*}

\begin{figure*}[!t]
\centering
\includegraphics[width=0.8\linewidth]{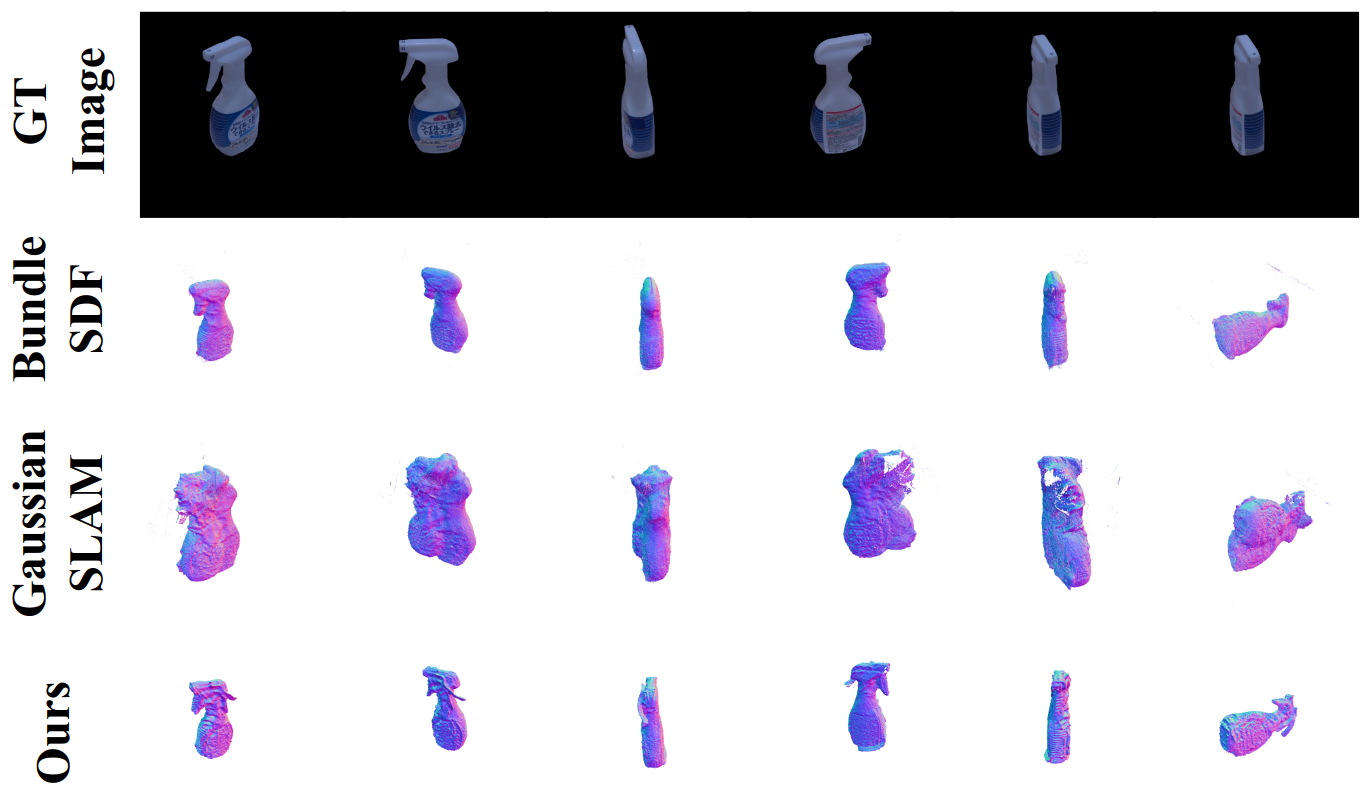}
\caption{Mesh Reconstruction Comparison on GTR3D Real, Spray Object 
}
\label{fig:spray_tsdf}
\end{figure*}

\begin{figure*}[!b]
\centering
\includegraphics[width=0.8\linewidth]{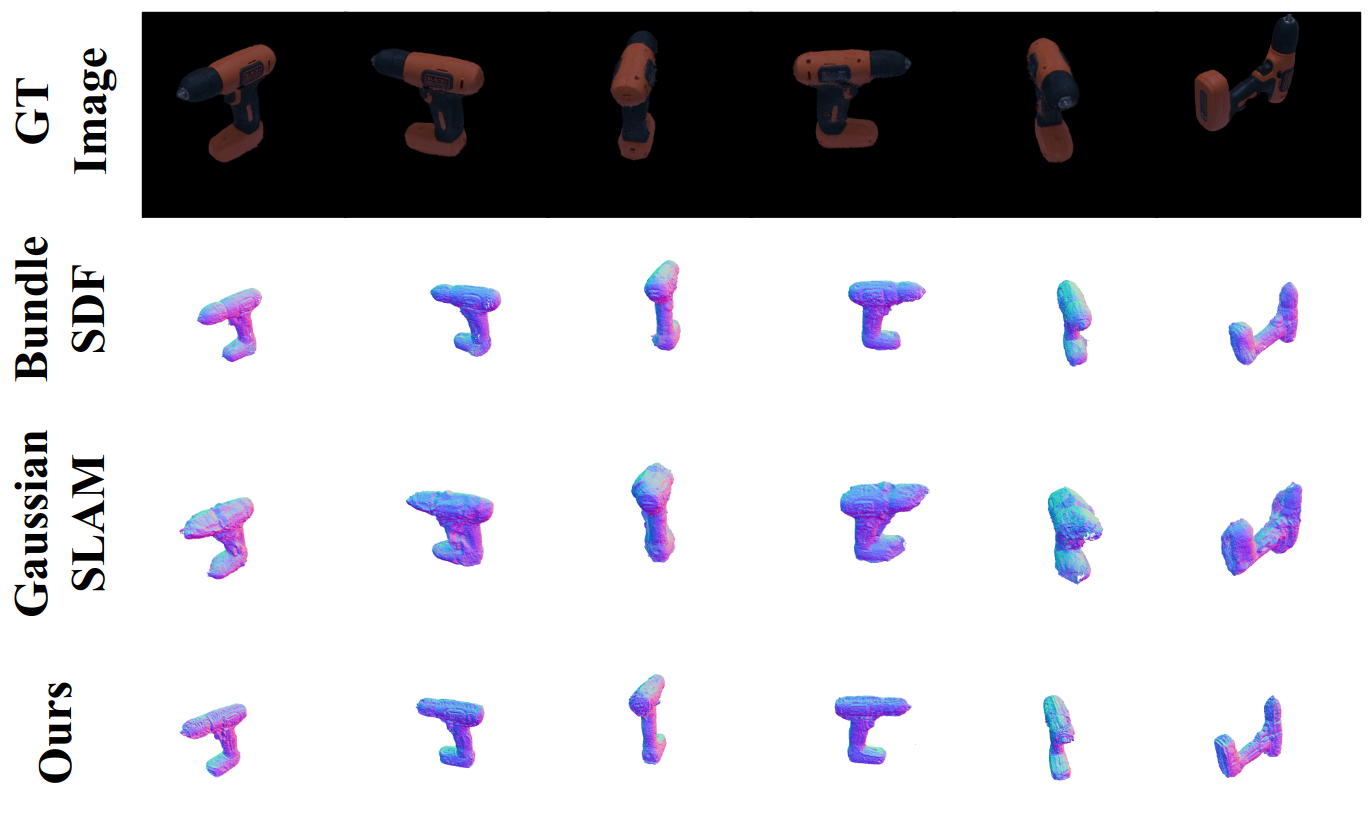}
\caption{Mesh Reconstruction Comparison on GTR3D Real, Drill Object 
}
\label{fig:drill_tsdf}
\end{figure*}

In \cref{fig:box_tsdf}, \cref{fig:can_tsdf}, \cref{fig:dino_tsdf}, \cref{fig:oil_tsdf}, \cref{fig:spray_tsdf}, \cref{fig:drill_tsdf} , we show reconstructed mesh via TSDF fusion described in Sec.3.4, from multiple viewpoints to observe full shape. % For BundleTrack and 
For BundleSDF, the back sides of the can and bottle, that are axis-symmetric objects, are missing since they fail to track the yaw movement for these objects as shown in \cref{fig:can_tsdf} and \cref{fig:oil_tsdf}. On the other hand, GaussianSLAM and our method are able to reconstruct the back side of these objects. Our reconstructed mesh is closer to the real one than GaussianSLAM. Only our approach can cover the full shape without collapse.

Furthermore, we show the point clouds that are iteratively fused based on our estimated pose in temporal order in ~\cref{fig:sup_teaser}.

% \section{Threshold of Appearance and Geometry Complexity}
% test

\section{Limitations and Future Work}

As we described in \cref{inf_time}, the computational time of 3DGS and PGO is much slower than that of other components. For 3DGS reconstruction, there is room for investigation into create 3DGS without optimization when the observed RGB-D image is of high quality, as the colored point cloud from RGB-D image are already available. 
This could potentially make the 3DGS reconstruction process much faster while maintaining the performance of render\&compare. 
For PGO, the CUDA implementation with an analytic Jacobian would make the process faster, although we utilized the pypose, a pytorch-based non linear optimization tool, for concise implementation. 

The investigation into additional material properties, such as specularity and transparency, is lacking, although we cover a wide variety of textures. The major challenge posed by these properties is appearance inconsistency over time.
% For example, the water is one of axial symmetry objects and contains transparency. 
For example, the appearance of the transparent areas of water bottle can vary based on the direction of light. In such cases, even dense appearance-based render\&comapre via 3DGS may struggle to track the yaw movement. 
To address this problem, we believe that performing render\&compare in feature space, rather than in RGB space, is a promising approach. 

% Lastly, we would like to decrease the parameters in the pipeline with keeping the performance. 

\section{Details of Baselines}

\textbf{Gaussian-SLAM~\cite{yugay2023gaussian}:} While Gaussian-SLAM is designed for a scene-level setting, we tuned its hyperparameters to make it work for an object-level tracking setup. Specifically, we found out a lower learning rate~(i.e. $8 \times 10^{-5}$ for rotation and $1 \times 10^{-5}$ for translation) produces the best results. To make it fair with our setup, we used mapping iterations of 1000 and tracking iterations of 500, similar to our setup. We further set the new point radius as $1 \times 10^{-7}$, alpha threshold 0.6 and pruning threshold 0.1 for mapping and set the color loss weight to 0.98 and depth loss weight to 0.02 for tracking. For a fair comparison, we disabled the camera initialization using constant speed or odometry supported in the implementation of Gaussian-SLAM, rather we initialize the camera using previously predicted camera pose, i.e. the relative difference between two is always identity. We also initiate a new submap every 4 frames. 

\textbf{FlowMap~\cite{smith24flowmap}:} As mentioned in the main paper, we considered FlowMap as an additional baseline to compare against, but excluded it due to the difference in settings~(i.e. sequential processing vs access to the full video sequence from the start) as well as the type of input used~(i.e. RGB-D used by our setting vs RGB-only used in FlowMap). To adapt FlowMap to our setting, we fixed the depth maps and camera intrinsic during the optimization procedure. This effectively reduces it to a non-learning based pipeline without any learnable parameters, since the only learnable parameters in FlowMap are the weights of a depth neural network and weights of the correspondence confidence MLP. With this adaptation, we solve for the rigid transformation between sequential pointclouds obtained by back projecting the depth maps using known camera intrinsic. For correspondences, we use off-the-shelf optical flow~\cite{teed2020raft} between sequential frames. This approach relies on the generalizability of the used optical flow model in a zero-shot manner. However, we found that in our object-centric setting, it struggles to find meaningful poses, with slipping observed from the very beginning, further emphasizing the challenging nature of our problem setting. 

% To split the supplementary pages from the main paper, you can use \href{https://support.apple.com/en-ca/guide/preview/prvw11793/mac#:~:text=Delete%20a%20page%20from%20a,or%20choose%20Edit%20%3E%20Delete).}{Preview (on macOS)}, \href{https://www.adobe.com/acrobat/how-to/delete-pages-from-pdf.html#:~:text=Choose%20%E2%80%9CTools%E2%80%9D%20%3E%20%E2%80%9COrganize,or%20pages%20from%20the%20file.}{Adobe Acrobat} (on all OSs), as well as \href{https://superuser.com/questions/517986/is-it-possible-to-delete-some-pages-of-a-pdf-document}{command line tools}.

%%%%%%%%%%%%%%%%%%%%%%%%%%%%%%%%%%%%%

